\documentclass{article}

    \PassOptionsToPackage{numbers, sort&compress}{natbib}

\usepackage[preprint]{neurips_2025}

\usepackage[utf8]{inputenc} 
\usepackage[T1]{fontenc}    
\usepackage{hyperref}       
\usepackage{url}            
\usepackage{booktabs}       
\usepackage{amsfonts}       
\usepackage{nicefrac}       
\usepackage{microtype}      
\usepackage{xcolor}         

\usepackage[utf8]{inputenc} 
\usepackage[T1]{fontenc}    
\usepackage{hyperref}       
\usepackage{url}            
\usepackage{booktabs}       
\usepackage{amsfonts}       
\usepackage{nicefrac}       
\usepackage{microtype}      
\usepackage{xcolor}         

\usepackage{graphicx}       
\usepackage{subcaption}     
\usepackage{amsmath}
\usepackage{multirow}
\usepackage{pgfplots}
\usepackage{pgfplotstable}
\usepackage{tikz}
\usepackage{filecontents}
\usepackage{comment }
\pgfplotsset{compat=1.18}
\usetikzlibrary{plotmarks}

\title{Any-Class Presence Likelihood for Robust Multi-Label Classification with Abundant Negative Data}

\author{%
\begin{tabular}[t]{c}
Dumindu Tissera$^{1}$, Omar Awadallah$^{1}$, Muhammad Umair Danish$^{1}$, \\
Ayan Sadhu$^{2}$, Katarina Grolinger$^{1}$
\end{tabular}\\
$^{1}$Department of Electrical and Computer Engineering, \\
$^{2}$Department of Civil and Environmental Engineering \\
Western University, London, Ontario, Canada \\
\texttt{\{gmigelhe, oawadall, mdanish3, asadhu, kgroling\}@uwo.ca}
}

\begin{document}

\maketitle

\begin{abstract}   
    Multi-label Classification (MLC) assigns an instance to one or more non-exclusive classes. A challenge arises when the dataset contains a large proportion of instances with no assigned class, referred to as negative data, which can overwhelm the learning process and hinder the accurate identification and classification of positive instances. Nevertheless, it is common in MLC applications such as industrial defect detection, agricultural disease identification, and healthcare diagnosis to encounter large amounts of negative data. Assigning a separate negative class to these instances further complicates the learning objective and introduces unnecessary redundancies. To address this challenge, we redesign standard MLC loss functions by deriving a likelihood of any class being present, formulated by a normalized weighted geometric mean of the predicted class probabilities. We introduce a regularization parameter that controls the relative contribution of the absent class probabilities to the any-class presence likelihood in positive instances. The any-class presence likelihood complements the multi-label learning by encouraging the network to become more aware of implicit positive instances and improve the label classification within those positive instances. Experiments on large-scale datasets with negative data: SewerML, modified COCO, and ChestX-ray14, across various networks and base loss functions show that our loss functions consistently improve MLC performance of their standard loss counterparts, achieving gains of up to 6.01 percentage points in F1, 8.06 in F2, and 3.11 in mean average precision, all without additional parameters or computational complexity. Code available at: \url{https://github.com/ML-for-Sensor-Data-Western/gmean-mlc}    
\end{abstract}
\section{Introduction}
    In Multi-label Classification (MLC), each instance can belong to several labels simultaneously, contrasting with multi-class classification, where each instance belongs to only one of several mutually exclusive classes. The ability to assign multiple non-exclusive labels makes MLC a more generic formulation, encompassing multi-class classification as a special case. MLC is widely applicable in real-world scenarios such as image tagging~\cite{lin2014microsoft, kuznetsova2020open}, genre classification~\cite{montalvo2023improving}, text categorization~\cite{chalkidis2019neural, schopf2024efficient}, industrial defect detection~\cite{hu2023toward, dang2022defecttr}, and disease diagnosis in medical applications~\cite{wang2017chestx, irvin2019chexpert, bien2018deep}.

    MLC solutions commonly treat each class as a separate output, evaluating whether each label is present or absent for a given instance. Explicitly adding an output neuron to indicate instances with no applicable labels would introduce unnecessary redundancy, since a target output of all zeros already naturally represents such cases. MLC faces unique challenges when the dataset contains many negative instances~\cite{haurum2021sewer, wang2017chestx} as these instances can dominate the training process and suppress the multi-label learning signal for positive cases, thereby hindering the model’s ability to accurately identify labels within positive instances. Real-world MLC scenarios with many negative data points are common: examples include defect classification with normal instances~\cite{haurum2021sewer}, disease classification with healthy instances~\cite{wang2017chestx, irvin2019chexpert}, and object recognition in specific contexts.

    Recent advances in loss function design, such as focal loss~\cite{lin2017focal} and class-balanced loss~\cite{cui2019class}, have shown promise in mitigating the impact of long-tail distributions and imbalanced data in generic classification, which also benefits MLC. At the same time, approaches that model label correlations, such as classifier chains~\cite{brinker2006active, BOGATINOVSKI2022117215} and graph-based methods~\cite{8961143}, enhance predictive performance by exploiting co-occurrence patterns among labels. Despite these developments, explicit modeling of the likelihood of any class being present in the presence of abundant negative data has been overlooked.

    We introduce a novel objective function that explicitly optimizes the model likelihood of any-class presence, alongside the standard MLC objective, to improve classification under overwhelming negative instances. The any-class presence likelihood is formulated as a normalized geometric mean over individual class probabilities, which encourages the model to contrast label classification in positive instances against negative ones, without introducing redundancy through extra output neurons or two-stage classification schemes for handling negative data. A regularization hyperparameter modulates the contribution of absent-class predictions to the any-class likelihood of positive instances, allowing the control of the learning signal. We integrate this objective into standard MLC losses, including Binary Cross Entropy and Focal Loss, and extend them with a class-balanced reweighting to address label imbalance and negative dominance. Extensive experiments across industrial defect detection, medical diagnosis, and object recognition tasks demonstrate consistent improvements over standard losses, without adding parameters or inference-time complexity.

\section{Related work} \label{sec:relatedwork}
     \paragraph{Traditional MLC techniques:} 
     MLC techniques have been successfully deployed in real-world applications such as medical diagnosis, content tagging, and industrial fault detection~\cite{durand2019learning, BOGATINOVSKI2022117215}. Binary Relevance (BR) is one of the earliest MLC methods, which transforms MLC into a set of independent binary classification tasks, each sharing the same feature space. While this method is intuitive, it is unsuitable for a large number of class labels and ignores label correlations~\cite{zhang2018binary, mao2024multilabel}. Calibrated Label Ranking (CLR)~\cite{brinker2006active} is another widely used method that converts MLC into a set of pairwise binary classifications; however, it is also not suitable for a large number of classes, due to the increased computational complexity associated with pairwise comparison~\cite{furnkranz2008multilabel}. Classifier Chains (CC)~\cite{read2011classifier} introduce label correlation but are limited by their reliance on the label ordering and face scalability issues similar to those of BR~\cite{read2011classifier}. Despite the successful deployments in real-world scenarios, BR, CLR, and CC do not address the problem of certain classes having very few training instances, making it difficult for models to learn those classes~\cite{furnkranz2008multilabel, read2011classifier, cole2021multi}. 

     \paragraph{Advanced MLC solutions:}
    In real-world multi-label datasets, negative instances often overwhelm the training signal~\cite{zhang2023deep, zhang2023noisy}, and the positive labels themselves frequently follow a long-tail distribution, where few labels occur very often, while most labels are rare~\cite{yuan2019long, wu2020distribution}. To address this, focal loss~\cite{deng2024edcloc} down-weights well-classified instances (typically abundant negatives) and focuses the model on hard, under-represented cases~\cite{lin2017focal, ridnik2021asymmetric}.  \citet{li2024semi} apply Binary Angular Margin Loss to balance the variance bias between label-wise positives and negatives. Class-balanced loss re-weights each label's contribution based on the \emph{effective number} of its positive instances, preserving the learning signal for tail classes~\cite{cui2019class}. Another alternative, post-hoc logit adjustment, shifts decision thresholds using estimated class priors, improving calibration for minority labels without additional sampling or re-weighting during training~\cite{menon2021longtail}. While these methods alleviate imbalance among positive labels~\cite{vu2024lcsl, gorishniy2021revisiting}, they treat each label independently and do not explicitly balance the overwhelming number of negative instances or leverage label co-occurrence patterns. Indeed, a large-scale empirical study across 42 benchmark datasets and 26 methods has shown that uncorrected long-tail distributions and unlabeled negatives substantially degrade MLC performance~\cite{BOGATINOVSKI2022117215, liu2021emerging}.

    Class co-occurrence has been studied and applied in various MLC applications~\cite{wang2024multi, rawlekar2025improving, wang2022multi, yu2024evidential, chen2024in}. For example, probabilistic techniques have been employed to estimate co-occurrence priors, which are then used to refine prediction scores without altering the model architecture~\cite{rawlekar2025improving}. Other ML-based methods, such as attention-based decoders~\cite{ridnik2021mldecoder, chen2024in} learn implicit class dependencies using a transformer-based classification head that decodes label predictions from a shared latent space, providing scalability across large datasets. Graph-based methods such as CheXGCN~\cite{8961143} utilize Graph Convolutional Networks to encode label relationships in medical imaging to enhance learning. Li et al.~\cite{li2024coocnet} proposed CoocNet for multi-label text classification, which improves label co-occurrence modeling. Additionally, contrastive learning techniques have been employed to enforce semantic similarity among co-occurring labels, enhancing the structure of the learned feature space~\cite{dao2021contrastive}. However, these techniques alter network architectures or add classifier heads, incurring additional compute.

    \paragraph{Gap in existing approaches:}
    In many real-world applications, datasets often contain a disproportionate number of negative instances~\cite{tarekegn2021review}. These negative instances can dominate the training process, making it difficult for models to learn positive instances~\cite{tarekegn2021review,sun2010multi, liu2021emerging}. For example, the ChestX-ray14 dataset~\cite{wang2017chestx} contains numerous normal scans that do not show disease symptoms. Similarly, agricultural disease detection datasets usually contain many healthy plant images that do not contribute directly to disease recognition~\cite{mohanty2016}. In industrial applications, normal products are often overrepresented, leading to considerable bias in model predictions~\cite{bergmann2019, tarekegn2021review}. Therefore, effective handling of such negative instances is essential to reduce bias and preserve model accuracy~\cite{huang2017multi, sun2010multi}.
    
    Despite considerable progress, current MLC methods either focus on class imbalance or label dependencies, but rarely both, while overlooking the overwhelming influence of negative instances commonly present in real-world applications. BR variants and imbalance‐aware losses (e.g., focal~\cite{lin2017focal} and class‐balanced loss~\cite{cui2019class}) neglect label co‐occurrence, whereas correlation models (e.g., classifier chains~\cite{read2011classifier}, GCNs~\cite{8961143}) do not explicitly counterbalance abundant negatives or long‐tail distributions. Consequently, no existing technique mitigates the dominance of fully negative data, rebalances imbalanced labels, and leverages label co‐occurrences, which motivates our any‐class presence likelihood. Moreover, we do this without adding parameters or altering the network architecture.

\section{Any-class presence likelihood} \label{sec:methods}
     To contrast the label classification in positive instances from negative instances, we propose optimizing the any-class presence likelihood as an explicit objective that complements the standard MLC class prediction optimization. We redesign commonly used objective functions in MLC: binary cross-entropy loss and the focal loss. Furthermore, a class-balanced weighting scheme is incorporated into the new loss function, ensuring balanced learning under imbalanced class distributions.

    \subsection{Redesigned binary cross-entropy with any-class presence likelihood}
    This subsection redesigns the standard Binary Cross-Entropy (BCE) by introducing any-class presence likelihood, derived directly from the predicted individual class probabilities. The neural network-based MLC is typically formulated as a task of predicting multiple binary labels, each assigned to a specific output unit. Assume an MLC task with \(M\) classes and a neural network \(f_\theta\) parameterized by \(\theta\). The network takes an instance \(\mathbf{x} \in \mathcal{X}\) as input and produces \(M\) raw scores, which are transformed into class prediction probabilities \(\mathbf{p} = (p_j)_{j=1}^M\) by the \(\mathrm{sigmoid}\) activation. The target is vector \(\mathbf{y} = (y_j)_{j=1}^M\), where \(y_j\) is 1 if class \(j\) is present and 0 otherwise. 
    The BCE loss for the instance is: 
    \begin{equation}
    \label{eq:j_bce}
        \mathcal{J}_{bce} = -\sum_{j=1}^{M} \log(p_j^t),
        \qquad \text{where} \qquad
        p_j^t =
        \begin{cases}
            p_j, & \text{if} \quad y_j = 1, \\
            1 - p_j, & \text{if} \quad y_j = 0.
        \end{cases}
    \end{equation} 
    The BCE loss can also be defined as the negative log of the model likelihood given \(\mathbf{x}\) and \(\mathbf{y}\):
    \begin{equation}
    \label{eq:l_bce}
        \mathcal{J}_{bce} = -\log(\mathcal{L}(\theta|\mathbf{x},\mathbf{y}), 
        \quad \text{where} \quad 
        \mathcal{L}(\theta|\mathbf{x},\mathbf{y}) = \prod_{j=1}^{M}p_j^t.
    \end{equation}
    Thus, a negative instance is implicitly represented by all (\(M\)) zero probabilities. However, BCE does not explicitly encourage the model to distinguish positive instance classification in the presence of a large number of negatives. We overcome this limitation by introducing an explicit any-class presence likelihood optimization that operates alongside the standard individual class predictions, compelling the neural network to contrast positive instance classification from negative instances.
    
    \paragraph{Any-class presence likelihood:} 
    For instance \(\mathbf{x}\), the target for any-class presence \(y_a\) is defined as 1 if any class is present, and 0 otherwise (i.e., all classes absent):
    \begin{equation}
        y_a =
        \begin{cases}
            1, & \text{if } \exists j \in \{1, \dots, M\} \text{ such that } y_j = 1 \\
            0, & \text{if } \forall j \in \{1, \dots, M\},\ y_j = 0
        \end{cases}
    \end{equation}
    We synthesize the predicted any-class presence probability \(p_a\) by taking the weighted geometric mean of the predicted class probabilities, normalized by the sum of the weighted geometric means of the predicted probabilities and their complements, as expressed in:
    \begin{equation}
    \label{eq:p_a}
        p_a = \frac{(\prod_{j=1}^{M}p_j^{w_j})^{1/\sum_{j}w_j}}{(\prod_{j=1}^{M}p_j^{w_j})^{1/\sum_{j}w_j} + (\prod_{j=1}^{M}(1-p_j)^{w_j})^{1/\sum_{j}w_j}}
        \quad \text{where} \quad
        w_j =
        \begin{cases}
        1, & \text{if} \quad y_j=1, \\
        \lambda, & \text{if} \quad y_j=0.
        \end{cases}
    \end{equation}
    Here, the weight \( w_j \) is associated with class \( j \) presence in the instance: it is set to 1 if class \( j \) is present and to a hyperparameter \( \lambda \in [0,1] \) if class \( j \) is absent, thereby adjusting the absent class contribution to the collective any-class presence probability. The geometric mean fairly represents central tendency~\cite{Vogel02012022} and maintains robustness and scale consistency compared to the product, as illustrated in Appendix~\ref{apx:likelihood}. The model likelihood for any-class presence is thus:
    \begin{equation}
    \label{eq:l_ya}
        \mathcal{L}(\theta|\mathbf{x},y_a) = p_a^t \quad \text{with} \quad p_a^t =
        \begin{cases}
        p_a, & \text{if} \quad y_a=1, \\
        1-p_a, & \text{if} \quad y_a=0.
        \end{cases}
    \end{equation}
    In the specific case of a negative instance (i.e., \( y_a = 0 \)), the any-class presence likelihood becomes:
    \begin{equation}
        \mathcal{L}(\theta|\mathbf{x},y_a=0) = 1 - p_a = \frac{ \left( \prod_{j=1}^{M} (1-p_j) \right)^{1/M} }{ \left( \prod_{j=1}^{M} p_j \right)^{1/M} + \left( \prod_{j=1}^{M} (1-p_j) \right)^{1/M} }.
    \end{equation}    
    In this case, the likelihood depends only on predicted probabilities of all classes and their complements, remaining unaffected by \(\lambda\). Therefore, \(\lambda\) only influences positive instances where at least one class is present. In positive instances, when \(\lambda = 0\), the any-class presence probability \(p_a\) is computed solely based on the predicted probabilities of present classes and their complements, using the normalized geometric mean. When \(\lambda > 0\), the absent classes are also included proportionally in the calculation, allowing their predicted probabilities to affect \(p_a\). This introduces a controlled amount of regularization into the optimization process, as the contribution of absent classes.

    \paragraph{Redesigned BCE with any-class likelihood:}    
    Having defined the any-class presence likelihood \(\mathcal{L}(\theta|\mathbf{x},y_a)\), the total model likelihood now becomes the product between the MLC likelihood (Equation~\eqref{eq:l_bce}) and the any-class presence awareness likelihood:
    \begin{equation}
    \label{eq:l_y_ya}
        \mathcal{L}(\theta|\mathbf{x}, \mathbf{y}, y_a) = \left( \prod_{j=1}^{M} p_j^t \right) (p_a^t)^\alpha,
    \end{equation}
    where \( \alpha \in [0,1] \) is the relative contribution of the any-class presence likelihood to the total likelihood.  Accordingly, the redesigned BCE loss is defined as the negative log of this total likelihood:
    \begin{equation}
    \label{eq:j_any_bce}
        \mathcal{J}_{any|bce} = -\log(\mathcal{L}(\theta|\mathbf{x}, \mathbf{y}, y_a)) = -\sum_{j=1}^{M} \log(p_j^t) - \alpha \log(p_a^t).
    \end{equation}
    This jointly optimizes the individual class predictions and the collective any-class presence likelihood.

    \subsection{Redesigned focal loss with any-class presence likelihood}
     
    We extend the formulated any-class presence likelihood to the focal loss, enabling the model to better handle long-tail distributions. In binary classification tasks, the abundance of easily detectable negative instances often dominates the learning process, making it difficult for the model to accurately learn from the hard-to-detect positive minority. Focal loss~\cite{lin2017focal} adaptively re-weights the cross-entropy objective by increasing the contribution of hard-to-classify predictions and reducing the contribution of correct predictions. Given a predicted probability vector \(\mathbf{p}\) and a target vector \(\mathbf{y}\), focal loss re-weights each cross-entropy term by the complement of the predicted confidence \(p_j^t\):
    \begin{equation}
    \label{eq:j_focal}
        \mathcal{J}_{focal} = -\sum_{j=1}^{M}(1-p_j^t)^\gamma \log(p_j^t).
    \end{equation}
    The original formulation of focal loss does not enforce learning of any-class presence; however, its adaptive re-weighting can benefit the any-class presence optimization by down-weighting easy negatives and up-weighting hard positives. Therefore, we reformulate the focal loss to incorporate the any-class presence likelihood, following a similar approach to BCE. We adaptively re-weight the log likelihood of any-class presence by the complement of the any-class presence confidence of \(p_a^t\):  
    \begin{equation}
    \label{eq:j_any_focal}
        \mathcal{J}_{any|focal} = -\sum_{j=1}^{M}(1-p_j^t)^\gamma \log(p_j^t)  - \alpha (1-p_a^t)^\gamma \log(p_a^t).
    \end{equation}
    This reformulated design enables the model to jointly optimize individual class predictions and the collective any-class presence signal under long-tail distributions.

    \subsection{Class-balanced reweighting for multi-label learning}

    To further improve learning from the imbalanced class distributions, we incorporate class-balanced weighting into the loss functions. Class-balanced loss~\cite{cui2019class} further weights each instance loss by the effective number of instances from the present classes. The effective number of instances \(n^e_j\) for each class \(j\) and class \(j\) balancing weight \(w^{cb}_j\)---which is the normalized \(n^e_j\)---are calculated as follows:
    \begin{equation}
        n^e_j = \frac{1-\beta}{1-\beta^{n_j}} \quad \text{and} \quad 
        w^{cb}_j = \frac{M n^e_j}{\sum_k n^e_k},
    \end{equation}
    where \(n_j\) is the number of instances in the dataset that have class \(j\) (\(y_j=1\)), and \(\beta \in [0,1)\) is a hyperparameter balancing between no weighting (\(\beta=0\)) and weighting by inverse class frequency (\(\beta \approx 1\)). Thus, the loss for each instance is weighted by the cumulative class-balancing weight, which is the sum of weights of all classes present in the instance. The class-balanced loss for the BCE objective from Equation~\eqref{eq:j_bce} becomes:    
    \begin{equation}
    \label{eq:j_bce_cb}
        \mathcal{J}^{cb}_{bce} = - \tilde{w}^{cb} \sum_{j=1}^{M} \log(p_j^t), \quad where \quad
        \tilde{w}^{cb} = \sum_{j \in y_j=1}w^{cb}_j
    \end{equation}
    We extend the class-balanced loss to accommodate the specific case with negative data by introducing a class-balancing weight for the negative instances. The negative instances in the dataset are counted, and the effective number of negative instances (\(n^e_{neg}\)) and the class-balancing weight for negative data (\(w^{cb}_{neg}\)) are calculated, considering the negative instances as an additional category. Without such a redesign, the negative instances get the class-balancing weight of zero. The normalization of the effective number of class instances to derive class-balancing weight is also adjusted.    
    \begin{equation}
        n^e_{neg} = \frac{1-\beta}{1-\beta^{n_{neg}}}, \quad \text{therefore} \quad 
        w^{cb}_j = \frac{(M+1)n^e_j}{\sum_k n^e_k + n^e_{neg}} \quad \text{and} \quad
        w^{cb}_{neg} = \frac{(M+1)n^e_{neg}}{\sum_k n^e_k + n^e_{neg}},
    \end{equation}
    where \(n_{neg}\) is the number of negative instances in the dataset (\(y_a=0\)). Now, the cumulative class-balancing weight for an instance becomes either the summation of balancing weights of the present classes if the instance is positive or the negative instance weight if it is negative:
    \begin{equation}
        \tilde{w}^{cb} = 
        \begin{cases} 
        \sum_{j \in y_j=1}w^{cb}_j, & if \quad y_a=1,\\
        w^{cb}_{neg}, & if \quad y_a=0.
        \end{cases}
    \end{equation}
    Accordingly, the redesigned BCE objective (Equation~\eqref{eq:j_any_bce}) is updated with class-balancing support:
    \begin{equation}
    \label{eq:j_any_bce_cb}
        \mathcal{J}^{cb}_{any|bce} = - \tilde{w}^{cb} \left[ \sum_{j=1}^{M} \log(p_j^t) - \alpha \log(p_a^t) \right].
    \end{equation}    
    Similarly, the redesigned focal loss (Equation~\eqref{eq:j_any_focal}) is updated with class-balancing support: 
    \begin{equation}
    \label{eq:j_any_focal_cb}
        \mathcal{J}^{cb}_{any|focal} = - \tilde{w}^{cb} \left[ \sum_{j=1}^{M} (1-p_j^t)^\gamma \log(p_j^t) - \alpha (1-p_a^t)^\gamma \log(p_a^t) \right].
    \end{equation}
    The reformulated objectives provide a principled approach to leverage both per-class and collective signals under an imbalanced multi-label setting. Appendix~\ref{apx:backprop} derives the learning signal on the final layer that encourages learning any-class presence collectively.

\section{Experiments} \label{sec:experiments}
    We evaluate our losses (\(\mathcal{J}^{cb}_{any|bce}\), \(\mathcal{J}^{cb}_{any|focal}\)) and their standard counterparts (\(\mathcal{J}^{cb}_{bce}\), \(\mathcal{J}^{cb}_{focal}\))  on large-scale image datasets representing distinct applications: SewerML~\cite{haurum2021sewer} for industrial defect detection, ChestX-ray14~\cite{wang2017chestx} for medical diagnosis, and modified COCO~\cite{lin2014microsoft} for object recognition. We also conduct an ablation study on the hyperparameter $\lambda$, which controls the weight of the absent classes in any-class presence probability calculation, to quantify its contribution.

    \subsection{Datasets}
    \textbf{SewerML} contains 1.3 million sewer-pipe images labeled across 17 defect types; of these, 1 million images form the training set and 130k each form the validation and test sets. In addition to the heavily imbalanced class distribution, nearly 53\% of the instances are normal, creating a strong negative-data presence. \textbf{ChestX-ray14} consists of 112k frontal-view X-rays from 31k patients, labelled across 14 pathologies. Approximately 60k (54\%) instances do not contain any considered pathologies. As the original dataset only provides train and validation data together with a separate test split, and the test split label distribution tremendously mismatches the train and validation~\cite{kufel2023multi}, we re-partition all images into 70\% train, 15\% validation, and 15\% test, ensuring that one patient appears only in one split. \textbf{COCO} is a benchmark for object detection, segmentation, and captioning, containing 80 object classes, and we use its instance classification annotations. Since COCO does not have negative instances, we re-label the dataset, keeping only a subset of the original labels. Since the \textit{person} class is the most frequent class, present in almost 60\% of the images, we remove the \textit{person} label and its most co-occurring class labels to create negative instances. Due to the unavailability of test set labels, we re-partition the 123k training and validation instances into a 70-15-15\% partitions for training, validation, and test. The modified COCO dataset has 41 classes; approximately 58k (47\%) instances are negative. Appendix~\ref{apx:coco-filtering} further explains the COCO label processing.

    \subsection{Models, training, and inference}
    We train three networks of varying architectural design: \textbf{TresNet-L}~\cite{ridnik2021tresnet}: a convolution-based network, \textbf{Vision Transformer (ViT-B16)}~\cite{dosovitskiy2021an}: an attention-based network, and \textbf{MaxViT-S}~\cite{tu2022maxvit}: a hybrid between convolutions and attention. TresNet-L is trained for 40 epochs with an SGD optimizer, a momentum of 0.9, batch size 256, and weight decay \(1\times10^{-4}\). The learning rate is set to 0.1 and decayed by a factor of 0.01 at epochs 20 and 30. ViT-B16 and MaxVit-S are trained for 90 epochs with an AdamW optimizer~\cite{loshchilov2018decoupled} (\(\beta_1=0.9\) and \(\beta_2=0.999\)), batch size 256, and a weight decay of 0.05. The learning rate is set to \(1\times10^{-3}\) for both networks (with an exception for MaxVit-S in ChestX-ray14, \(1\times10^{-4}\), due to difficulty in learning from chest data). The learning rate is varied by a cyclic decay with cosine annealing and warm restarts~\cite{loshchilov2017sgdr}, where the first cycle length \(T_0\) is 10 epochs, and each subsequent cycle length is multiplied by \(T_{mult}=2\). We also use a linear learning rate warm-up for five epochs, starting from \(1\times10^{-5}\). These hyperparameter selections are motivated by previous relevant work \cite{haurum2022multi, tu2022maxvit}. We set \(\alpha\) in our redesigned loss to 1 to reflect equal contribution of any-class presence likelihood to individual class likelihoods. Hyperparameter \(\lambda\) is set to 0.02, following our ablation experiment. The class balancing \(\beta\) is set to 0.9999, and focal \(\gamma\) is set to 2.

    We use several metrics, averaged across classes, to measure performance. The predicted presence of a class is derived by applying a threshold of 0.5 to its predicted probability. \textbf{F1}, the harmonic mean of precision and recall, is used as it is commonly applied with unbalanced datasets. \textbf{F2}, which emphasizes recall, is more useful with many negative instances, where false negatives are more critical than false positives. SewerML authors introduced the F2 Class Importance Weighting \textbf{F2-CIW}, which weights each class F2 by industry-set weights. Mean Average Precision \textbf{mAP}, which is threshold-agnostic, quantifies the ranking quality of predicted label probabilities. Due to its threshold-agnostic nature, we utilize mAP as the validation metric to optimize during training. Apart from measuring MLC performance, \textbf{F1-Neg}, the F1 of predicting negative instances is also reported.   

    \subsection{Results}
    \label{ss:results}

    The experiments conducted on the three datasets show that the models trained with our redesigned loss functions (\(\mathcal{J}^{cb}_{any|bce}\), \(\mathcal{J}^{cb}_{any|focal}\)) consistently outperform their standard counterparts (\(\mathcal{J}^{cb}_{bce}\), \(\mathcal{J}^{cb}_{focal}\)) in terms the label classification metrics (F1, F2, mAP) while also maintaining performance on negative instance detection (F1-Neg). On \textbf{SewerML} dataset, as shown in Table~\ref{tab:sewer_ml}, the redesigned loss variants consistently improved MLC performance over their standard counterparts across all network architectures as evident across multiple metrics: F1, F2-CIW, and mAP. For example, with TresNet-L, the validation F2-CIW score improves from 61.96 to 65.1 with the redesigned BCE loss, and from 57.66 to 64.22 with the redesigned focal loss. Similarly, for ViT-B16, the validation F2-CIW score rises from 45.67 to 52.61 with the redesigned BCE, and from 51.28 to 54.58 with the redesigned focal loss. MaxViT-S achieves the strongest performance with standard losses among the tested models, and the redesigned losses improve it further, raising the validation F1 score from 66.18 to 67.45 and F2-CIW from 66.51 to 69.04 with redesigned focal loss, and increasing mAP from 69.7  to 70.23 with the redesigned BCE loss. The labels for the SewerML test set are not publicly available, but we submitted our model predictions for the test set to the SewerML authors to obtain the test set metrics F2-CIW and F1-Neg. The trends observed on the test set mirror the validation performance, further confirming the consistent benefit of the redesigned losses.
    \begin{table}[h]
        \caption{SewerML: performance metrics. The test benchmark only provided F2-CIW and F1-Neg}
        \vspace{0.5\baselineskip}
        \label{tab:sewer_ml}
        \centering
        \begin{tabular}{lllllllll}
            \toprule
            \multirow{2}{*}{Model} & \multirow{2}{*}{Loss} & \multicolumn{4}{c}{Validation} & \multicolumn{2}{c}{Test} \\
            \cmidrule(lr){3-6} \cmidrule(lr){7-8}
            & &                                         \small{F1} & \small{F2-CIW} & \small{mAP}  & \small{F1-Neg} & \small{F2-CIW} & \small{F1-Neg} \\ 
            \midrule
            TresNet-L  & \(\mathcal{J}^{cb}_{bce}\)        & 63.13 & 61.96 & 65.63 & 91.9 & 60.02 & \textbf{91.58} \\ 
            (53.6M)    & \(\mathcal{J}^{cb}_{any|bce}\)    & \textbf{64.23} & \textbf{65.1} & \textbf{66.86} & \textbf{91.95} & \textbf{64.16} & 91.35 \\
                       & \(\mathcal{J}^{cb}_{focal}\)      & 59.73 & 57.66 & 62.71 & 91.21  & 56.08 & 90.99  \\
                       & \(\mathcal{J}^{cb}_{any|focal}\)  & 63.33 & 64.22 & 65.82 & 91.59  & 62.99 & 91.05 \\
           \midrule
            ViT-B16    & \(\mathcal{J}^{cb}_{bce}\)        & 48.25 & 45.67 & 52    & 89.55 & 44.39 & 88.85  \\ 
            (85.8M)    & \(\mathcal{J}^{cb}_{any|bce}\)    & 52.03 & 52.61 & 53.15 & 89.62 & 50.49 & 88.9  \\
                       & \(\mathcal{J}^{cb}_{focal}\)      & 51.85 & 51.28 & 54.09 & \textbf{89.95} & 48.6 & \textbf{89.44} \\
                       & \(\mathcal{J}^{cb}_{any|focal}\)  & \textbf{52.63} & \textbf{54.58} & \textbf{54.35} & 89.14 & \textbf{52.25} & 88.35 \\
           \midrule 
            MaxViT-S   & \(\mathcal{J}^{cb}_{bce}\)         & 66.99  & 67.15 & 69.47  & \textbf{93.34} & 65.64 & \textbf{92.83} \\  
            (68.2M)    & \(\mathcal{J}^{cb}_{any|bce}\)    & 67.05 & 68.16 & \textbf{70.23} & 92.96 & 66.73 & 92.46 \\
                       & \(\mathcal{J}^{cb}_{focal}\)       & 66.18 & 66.51 & 69.42 & 92.96 & 65.28 & 92.48 \\
                       & \(\mathcal{J}^{cb}_{any|focal}\)  & \textbf{67.45} & \textbf{69.04} & 70.18 & 93.15 & \textbf{67.41} & 92.58 \\
            \bottomrule
        \end{tabular}
    \end{table}
    
    In each scenario, F1, F2, and mAP improve from standard loss to redesigned loss, while also maintaining the F1-Neg, the F1 score of negative instance detection. While a drop in F1-Neg is expected with improved capture of positive instances, the decrease in F1-Neg is minimal (or nonexistent) compared to the gain in overall MLC performance. In MLC applications with many negative instances (e.g., defect or disease detection), it is often more important to reduce false negatives than to lower false positives. Our redesigned loss functions encourage the models to better learn the classification by contrasting the positive class predictions with all negative instances, reducing false negatives. For the remaining datasets, we only show the performance of the TresNet-L and MaxViT-S since ViT-B16 performed poorly irrelevant of the loss function used.

    \textbf{ChestX-ray14}, Table~\ref{tab:chest}, is a challenging dataset, exhibiting lower overall classification performance than the other two datasets. This is partly due to known label noise introduced by automated NLP-based annotations \cite{oakden2017exploring}. Nevertheless, the improvements in F-scores with redesigned losses are evident. For example, with TresNet-L, the F2 score improves from 12.77 to 20.43 on the validation split, and from 13.02 to 21.06 on the test split, when using the redesigned BCE. Similarly, MaxViT-S sees its F2 validation score increase from 16.58 to 23.22 and its F2 test score from 17.83 to 25.27 when using the redesigned focal loss. MaxViT-S with redesigned focal loss achieves the best overall test F-scores. This improvement in F-scores comes with minimal compromise in negative instance detection (F1-Neg), indicating substantially better classification of positive instances.

    \begin{table}
        \caption{ChestX-ray14 performance metrics}
        \vspace{0.5\baselineskip}
        \label{tab:chest}
        \centering
        \begin{tabular}{llllllllll}
            \toprule
            \multirow{2}{*}{Model} & \multirow{2}{*}{Loss} 
            & \multicolumn{4}{c}{Validation} 
            & \multicolumn{4}{c}{Test} \\
            \cmidrule(lr){3-6} \cmidrule(lr){7-10}
            & & \small{F1} & \small{F2} & \small{mAP} & \small{F1-Neg} 
              & \small{F1} & \small{F2} & \small{mAP} & \small{F1-Neg} \\
            \midrule
            TresNet-L  & \(\mathcal{J}^{cb}_{bce}\)       & 13.98   & 12.77  & 17.55  & \textbf{72.83}  & 14.63  & 13.02  & 18.57  & \textbf{72.45} \\  
            (53.6M)    & \(\mathcal{J}^{cb}_{any|bce}\)   & \textbf{19.99}   & \textbf{20.43}  & \textbf{18.53}  & 71.95  & \textbf{20.75} & \textbf{21.06}  & \textbf{19.67}  & 71.63  \\
                       & \(\mathcal{J}^{cb}_{focal}\)     & 8.17   & 7.03  & 15.23  & 72.24  & 9.21  & 8.12  & 15.86  & 71.81  \\
                       & \(\mathcal{J}^{cb}_{any|focal}\) & 13.68   & 13.83  & 15.69  & 72.29  & 14.19  & 14.26  & 16.63  & 71.72  \\
            \midrule 
            MaxViT-S   & \(\mathcal{J}^{cb}_{bce}\)       & 19.08  & 18.53  & \textbf{19.72}  & 72.85  & 20.7  & 19.78  & \textbf{21.26}  & 72.3  \\  
            (68.2M)    & \(\mathcal{J}^{cb}_{any|bce}\)   & \textbf{22.5}  & \textbf{23.99}  & 19.17  &  71.88  & 23.95  & 25.15  & 20.76  & 71.3  \\
                       & \(\mathcal{J}^{cb}_{focal}\)     & 17.41  & 16.58  & 18.28  & \textbf{73.48}  & 18.96  & 17.83  & 20.32  & \textbf{72.52}  \\
                       & \(\mathcal{J}^{cb}_{any|focal}\) & 22.14  & 23.22  & 18.99  & 71.98  & \textbf{24.46} & \textbf{25.27}  & 20.41  & 71.51  \\
            \bottomrule
        \end{tabular}
    \end{table}

    The \textbf{COCO} dataset, as seen from Table~\ref{tab:coco}, demonstrates similar results with our redesigned loss functions consistently improving both F-scores and mAP compared to the standard counterparts. For example, with TresNet-L, the redesigned BCE loss raises the validation F2 score from 36.41 to 41.59, and with MaxViT-S, the redesigned focal loss increases the validation F2 score from 40.47 to 46.85. The test split follows the same pattern. Notably, MaxViT-S trained with the redesigned focal loss achieves the highest overall MLC performance across both validation and test splits, with validation mAP reaching 46.78 and F1 score reaching 46.97.

    \begin{table}
            \caption{COCO perforamnce metrics}
            \vspace{0.5\baselineskip}
            \label{tab:coco}
            \centering
            \begin{tabular}{llllllllll}
                \toprule
                \multirow{2}{*}{Model} & \multirow{2}{*}{Loss} 
                & \multicolumn{4}{c}{Validation} 
                & \multicolumn{4}{c}{Test} \\
                \cmidrule(lr){3-6} \cmidrule(lr){7-10}
                & & \small{F1} & \small{F2} & \small{mAP} & \small{F1-Neg} 
                  & \small{F1} & \small{F2} & \small{mAP} & \small{F1-Neg} \\
                \midrule
                TresNet-L  & \(\mathcal{J}^{cb}_{bce}\)       & 40.93  & 36.41  & 42.52  & \textbf{77.58}  & 40.01  & 35.68  & 42.33  & \textbf{77.66}  \\ 
                (53.6M)    & \(\mathcal{J}^{cb}_{any|bce}\)   & \textbf{43.85}  & \textbf{41.59}  & \textbf{44.34}  & 76.58  & \textbf{42.92}  & \textbf{40.81} & \textbf{43.77}  & 76.52  \\
                           & \(\mathcal{J}^{cb}_{focal}\)     & 35.35  & 30.66  & 38.29  & 75.45  & 35.12  & 30.63  & 38.09  & 75.47  \\
                           & \(\mathcal{J}^{cb}_{any|focal}\) & 40.91  & 38.62  & 40.29  & 73.83  & 39.45  & 37.43  & 39.54  & 74.03  \\
                \midrule 
                MaxViT-S   & \(\mathcal{J}^{cb}_{bce}\)       & 38.52  & 34.65  & 40.56  & 76.77  & 37.88  & 34.43 & 40.24 & 77.09  \\ 
                (68.2M)    & \(\mathcal{J}^{cb}_{any|bce}\)   & 43.75  & 42.71  & 43.54  & 76.33  & 43.21 & 42.25  & 43.19  & 76.24  \\ 
                           & \(\mathcal{J}^{cb}_{focal}\)     & 43.46  & 40.47  & 44.8   & \textbf{79.62}  & 43.14  & 40.42  & 44.82  & \textbf{79.75}  \\
                           & \(\mathcal{J}^{cb}_{any|focal}\) & \textbf{46.97}  & \textbf{46.85}  & \textbf{46.78}  & 76.34  & \textbf{46.17}  & \textbf{45.97}  & \textbf{46.28}  & 76.69  \\
                \bottomrule
            \end{tabular}
    \end{table}
    
    These consistent improvements across datasets and networks confirm that the proposed modeling of any-class presence likelihood along standard MLC learning complements per-class objectives and mitigates the overwhelming effect of negative instances on positive instance classification.

    \subsection{Impact of \texorpdfstring{\(\lambda\)}{lambda}: the absent class contribution to the positive instances}
    \label{ss:ablation_lambda}

\begin{figure}[htbp]
        \centering
        \begin{subfigure}[b]{0.49\textwidth}
            \centering
            \pgfplotstableread{
lambda   F1     F2       AP     Recall  F1-Any  F1-Neg
0.005	63.85	63.23	66.19	63.04	91.31	91.77   
0.01	64.38	64.05	66.86	64.05	91.46	91.85
0.02	64.23	63.92	66.86	63.95	91.57	91.95
0.05	64.3	64.1	66.81	64.2	91.53	91.91
0.1	    63.2	62.96	65.55	63.04	91.34	91.75
0.2	    63.6	62.95	65.77	62.77	91.61	92.15
0.5	    63.07	61.73	65.06	61.11	91.39	92.13
1	    61.34	59.21	63.16	58.1	90.83	91.73
}\DataTableSewerTresBCE

\begin{tikzpicture}
\begin{loglogaxis}[
    xlabel={\scriptsize{$\lambda$ (in log scale)}},
    ylabel={\scriptsize{Score (\%)}},
    log ticks with fixed point,
    xtick={0.005, 0.01, 0.02, 0.05, 0.1, 0.2, 0.5, 1},
    xticklabels={0, 0.01, 0.02, 0.05, 0.1, 0.2, 0.5, 1},  
    xticklabel style={font=\scriptsize},
    ymin=55,
    ytick={55, 60, 65, 70, 75},
    yticklabels={55, 60, 65, 90, 95},  
    yticklabel style={font=\scriptsize, /pgf/number format/fixed},
    legend style={
        at={(0.5,-0.25)},
        anchor=north,
        draw=none,
        fill=none,
        font=\scriptsize,
        legend columns=5,
        /tikz/every even column/.style={column sep=0.3cm}
    },
    width=\linewidth,
    height=5cm,
    grid=major,
    xmajorgrids=false,
    ymajorgrids=true,
    tick style={black},
    every axis label/.append style={font=\small},
    label style={font=\small},
    title style={font=\footnotesize},
    legend cell align=left
]

\addplot+[color=blue, mark=o, thick, forget plot] table[x=lambda, y=F1] from \DataTableSewerTresBCE;
\addplot+[color=orange, mark=square, thick, forget plot] table[x=lambda, y=F2] from \DataTableSewerTresBCE;
\addplot+[color=purple, mark=triangle, thick, forget plot] table[x=lambda, y=AP] from \DataTableSewerTresBCE;
\addplot+[color=red!70!black, mark=star, thick, solid, forget plot] 
    table[x=lambda, y expr=\thisrow{F1-Neg}-20] from \DataTableSewerTresBCE;

\addplot+[color=blue, mark=o, only marks, thick] coordinates {(0.0025, 63.13)};
\addplot+[color=orange, mark=square, only marks, thick] coordinates {(0.0025, 60.82)};
\addplot+[color=purple, mark=triangle, only marks, thick] coordinates {(0.0025, 65.63)};
\addplot+[color=red!70!black, mark=star, only marks, thick] coordinates {(0.0025, 91.9 - 20)};

\path[fill=gray!30, fill opacity=0.5]
    (axis cs:0.0016, 55) rectangle (axis cs:0.0037, 90);

\node[font=\tiny\itshape] at (axis cs:0.0025, 57) {$\mathcal{J}^{cb}_{bce}$};
\node[font=\tiny\itshape] at (axis cs:0.01, 57) {$\mathcal{J}^{cb}_{any|bce}$};

\draw[black, thick, dashed] 
    (axis cs:0.0015, 68.5) -- (axis cs:1,68.5);

\addlegendimage{only marks, mark=o, color=blue}
\addlegendentry{F1}
\addlegendimage{only marks, mark=square, color=orange}
\addlegendentry{F2}
\addlegendimage{only marks, mark=triangle, color=purple}
\addlegendentry{AP}
\addlegendimage{only marks, mark=star, color=red!70!black}
\addlegendentry{F1-Neg}

\end{loglogaxis}
\end{tikzpicture}
            \caption{SewerML validation set}
            \label{fig:lambda-sewer-tresnet-bce}
        \end{subfigure}
        \hfill
        \begin{subfigure}[b]{0.49\textwidth}
            \centering
            \pgfplotstableread{
lambda   F1     F2     AP     Recall  F1-Any  F1-Neg
0.005       43.58	40.97	43.71	39.52	77.38	76.19
0.01    44.06	41.66	44.38	40.35	77.41	75.84
0.02    43.85	41.59	44.34	40.36	78.12	76.58
0.05    43.14	40.56	43.72	39.18	77.5	77.49
0.1     42.2	39.31	42.96	37.79	76.86	78.02
0.2     40.99	37.67	42.12	35.95	75.12	77.83
0.5     40.38	36.41	42.19	34.33	73.66	78.01
1       40.87	36.41	43.13	34.09	72.87	77.78
}\datatable

\begin{tikzpicture}
\begin{loglogaxis}[
    xlabel={\scriptsize{$\lambda$ (in log scale)}},
    ylabel={\scriptsize{Score (\%)}},
    log ticks with fixed point,
    xtick={0.005, 0.01, 0.02, 0.05, 0.1, 0.2, 0.5, 1},
    xticklabels={0, 0.01, 0.02, 0.05, 0.1, 0.2, 0.5, 1},  
    xticklabel style={font=\scriptsize},
    ymin=30,
    ytick={30, 35, 40, 45, 50, 55, 60},
    yticklabels={30, 35, 40, 45, 70, 75, 80},  
    yticklabel style={font=\scriptsize, /pgf/number format/fixed},
    legend style={
        at={(0.5,-0.25)},
        anchor=north,
        draw=none,
        fill=none,
        font=\scriptsize,
        legend columns=5,
        /tikz/every even column/.style={column sep=0.3cm}
    },
    width=\linewidth,
    height=5cm,
    grid=major,
    xmajorgrids=false,
    ymajorgrids=true,
    tick style={black},
    every axis label/.append style={font=\small},
    label style={font=\small},
    title style={font=\footnotesize},
    legend cell align=left
]

\addplot+[color=blue, mark=o, thick, forget plot] table[x=lambda, y=F1] from \datatable; 
\addplot+[color=orange, mark=square, thick, forget plot] table[x=lambda, y=F2] from \datatable; 
\addplot+[color=purple, mark=triangle, thick, forget plot] table[x=lambda, y=AP] from \datatable;
\addplot+[color=red!70!black, mark=star, thick, forget plot] table[x=lambda, y expr=\thisrow{F1-Neg}-20] from \datatable;

\addplot+[color=blue, mark=o, only marks, thick] coordinates {(0.0025, 40.93)};
\addplot+[color=orange, mark=square, only marks, thick] coordinates {(0.0025, 36.41)};
\addplot+[color=purple, mark=triangle, only marks, thick] coordinates {(0.0025, 42.52)};
\addplot+[color=red!70!black, mark=star, only marks, thick] coordinates {(0.0025, 77.58 - 20)};

\path[fill=gray!30, fill opacity=0.5]
    (axis cs:0.0016, 30) rectangle (axis cs:0.0037, 90);

\node[font=\tiny\itshape] at (axis cs:0.0025, 32) {$\mathcal{J}^{cb}_{bce}$};
\node[font=\tiny\itshape] at (axis cs:0.01, 32) {$\mathcal{J}^{cb}_{any|bce}$};

\draw[black, thick, dashed] 
    (axis cs:0.0015, 50) -- (axis cs:1,50);

\addlegendimage{only marks, mark=o, color=blue}
\addlegendentry{F1}
\addlegendimage{only marks, mark=square, color=orange}
\addlegendentry{F2}
\addlegendimage{only marks, mark=triangle, color=purple}
\addlegendentry{AP}
\addlegendimage{only marks, mark=star, color=red!70!black}
\addlegendentry{F1-Neg}

\end{loglogaxis}
\end{tikzpicture}
            \caption{COCO validation set}
            \label{fig:lambda-coco-tresnet-bce}
        \end{subfigure}
        \caption{Performance across varying $\lambda$ for TresNet-L in SewerML and COCO validation splits with redesigned BCE loss. The Y-axis is disjoint to remove idle plotting space between the F1-Neg curve and other curves. The metric values for standard BCE loss are plotted on the left of each plot.}
        \label{fig:lambda_abltation}
    \end{figure}
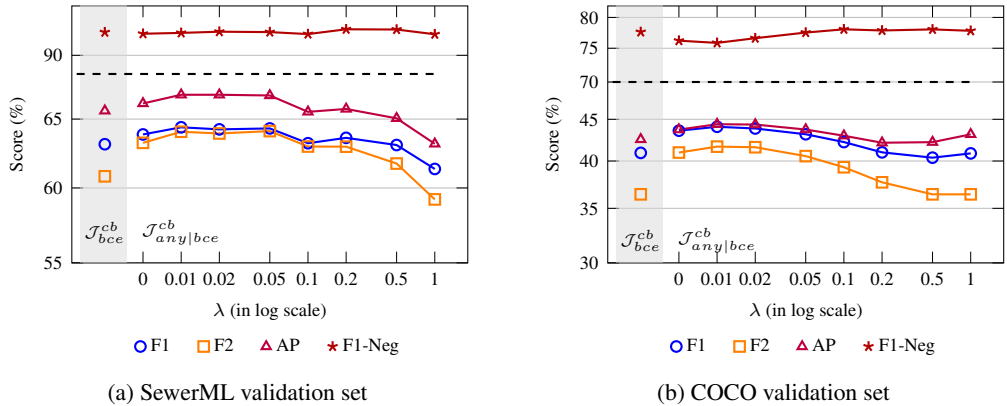

    We vary $\lambda$ to assess the impact of absent-class contribution to the any-class presence likelihood in positive instances. Table~\ref{tab:lambda_ablation} and Figure~\ref{fig:lambda_abltation} illustrate the results of this study for TresNet-L with redesigned BCE loss, evaluated on \textbf{SewerML} and \textbf{COCO} datasets. We report F1, F2, mAP, and F1-Neg of standard BCE loss \(\mathcal{J}^{cb}_{bce}\) and its redesigned counterpart \(\mathcal{J}^{cb}_{any|bce}\) for \(\lambda\) values between 0 and 1, with denser sampling between 0 and 0.1. Small $\lambda$ values, such as 0.01 and 0.02, yield the higher F-scores and mAP values, indicating that incorporating a small negative-class "\textit{regularization}" benefits more the geometric-mean aggregation in positive instances. For example, on the SewerML dataset, F1 improves from 63.13 with standard BCE to 64.38 with $\lambda=0.01$, while F2 increases from 60.82 to 64.05. On \textbf{COCO} dataset, peak performance occurs at $\lambda=0.01$ for both F1 (44.06) and mAP (44.38), compared to 40.93 and 42.52, respectively, with standard BCE. Setting $\lambda$ too large dilutes the any-class signal, with performance gradually declining as $\lambda$ approaches 1.0. Ablation on ChestX-ray14 with BCE loss and SewerML with focal loss (Appendix~\ref{apx:experiments}) shows similar results.

    \begin{table}[h]
    \setlength{\tabcolsep}{5.5pt}
    \caption{TresNet-L model performance with varying \(\lambda\)}
    \vspace{0.5\baselineskip}
    \label{tab:lambda_ablation}
    \centering
    \begin{tabular}{lllllllllll}
    \toprule
    \multirow{2}{*}{Dataset} &  \multirow{2}{*}{Metric} & \multirow{2}{*}{\(\mathcal{J}^{cb}_{bce}\)} & \multicolumn{8}{c}{ \(\mathcal{J}^{cb}_{any|bce}\) with varying \(\lambda\) }  \\
    \cmidrule(r){4-11}
    & & &  0  & 0.01 & 0.02  & 0.05  & 0.1  & 0.2  & 0.5 & 1  \\
    \midrule
    SewerML & F1 & 63.13   & 63.85  & \textbf{64.38} & 64.23  & 64.30  & 63.20  & 63.60  & 63.07 & 61.34 \\
            & F2 & 60.82   & 63.23  & 64.05 & 63.92  & \textbf{64.10}  & 62.96  & 62.95  & 61.73 & 59.21 \\
            & mAP & 65.63   & 66.19  & \textbf{66.86} & \textbf{66.86}  & 66.81  & 65.55  & 65.77  & 65.06 & 63.16 \\
            & F1-Neg & 91.9  & 91.77  & 91.85  & 91.95  & 91.91  & 91.75  & \textbf{92.15}  & 92.13  & 91.73\\     
    \midrule
    COCO   & F1 & 40.93    & 43.58  & \textbf{44.06}  & 43.85  & 43.14  & 42.2  & 40.99  & 40.38  & 40.87\\
             & F2 & 36.41    & 40.97  & \textbf{41.66}  & 41.59  & 40.56  & 39.31  & 37.67  & 36.41  & 36.41  \\
             & mAP & 42.52    & 43.71  & \textbf{44.38}  & 44.34  & 43.72  & 42.96  & 42.12  & 42.19  & 43.13  \\
             & F1-Neg & 77.58 & 76.19  & 75.84  & 76.58  & 77.49  & \textbf{78.02}  & 77.83  & 78.01  & 77.78 \\
    \bottomrule
    \end{tabular}
    \end{table}

    Across both datasets, F1-Neg is slightly compromised with redesigned loss for smaller \(\lambda\) (0 to 0.02) and improves with moderate values ($\lambda=0.2$), reflecting the tradeoff between classification within positive instances and detecting negative instances. Considering the context of enhancing classification of positive instances while maintaining awareness of negatives, we recommend setting \(\lambda\) in the range \([0.01,0.05]\) to balance between leveraging absent-class information and preserving the focus on present labels in positive instances. Overall, this ablation study confirms our hypothesis that a controlled contribution from absent classes to the any-class presence likelihood helps the model better contrast positive instance classification.

\section{Conclusion} \label{sec:conclusion}
    We proposed a principled extension to standard multi-label classification by explicitly optimizing the any-class presence likelihood, defined as the normalized geometric mean of individual class prediction probabilities. Experiments across diverse datasets and architectures demonstrated that our redesigned loss functions effectively address the challenge of overwhelmingly negative instances common in real-world MLC tasks, consistently improving per-class performance while preserving negative instance recognition. Ablation studies confirmed that a contribution from absent classes to the any-class presence likelihood enhances the classification.
    
    \textbf{Limitations and future work:} 
    The improved MLC performance comes with an expected trade-off: a slight reduction in negative instance detection in some cases, which is justifiable given the gains in MLC performance and the demands of the application contexts. Hyperparameter \(\lambda\) enables control over this trade-off, allowing us to balance optimal label classification with accurate negative instance detection. Furthermore, the proposed loss can be extended to settings with weak MLC annotations or cases where labels exhibit ambiguous presence/absence status. Another direction is to explore discrete classification heads for each class, enabling each class detector to be an expert in the individual class while using the any-class presence likelihood to maintain collective awareness of positive instances.

\bibliographystyle{unsrtnat}
\small
\bibliography{references}

\normalsize
\appendix

\section{Appendix / supplemental material}
\subsection{Learning signal on final layer neurons}
\label{apx:backprop}
In this Appendix, we explain how the redesigned loss function influences the learning behavior of the final layer neurons during training. We show that each output neuron receives the error between the synthesized any-class prediction and its target, in addition to the standard gradient during backpropagation, demonstrating how the proposed change contrasts the MLC classification with the negatives. To build this understanding, we first revisit the formulation of conventional MLC with BCE, from input to loss calculation, then break down how each final layer neuron updates based on the error. The network takes an instance \(\mathbf{x} \in \mathcal{X}\) as input and the final layer neurons produce \(M\) raw scores \(\mathbf{Z} = (z_j)_{j=1}^M\). The \(\mathrm{sigmoid}\) activation transforms raw scores into the class prediction probabilities \(\mathbf{p} = (p_j)_{j=1}^M\), where 
\begin{equation}
\nonumber
    p_j = \frac{1}{1 + e^{-z_j}}. \quad \text{Which also means that} \quad e^{-z_j} = \frac{1 - p_j}{p_j} \quad \text{(for later derivations).}
\end{equation}
For simplicity of explanation, the BCE loss from Equation~\eqref{eq:j_bce} is rewritten as: 
\begin{equation}
   \mathcal{J}_{bce} = -\sum_{j=1}^{M}[( y_jlog(p_j) + (1-y_j)log(1-p_j)].
\end{equation}
and the redesigned BCE loss from Equation~\eqref{eq:j_any_bce} as:
\begin{equation}
    \mathcal{J}_{any|bce} = \mathcal{J}_{bce} + \alpha \mathcal{J}_{any} \quad \text{where} \quad
    \mathcal{J}_{any} = - y_a log(p_a) - (1 - y_a) log(1 - p_a).
\end{equation}

Therefore, the gradient propagated to each final layer neuron \(z_j\) can be formulated as follows;
\begin{equation}
    \frac{\partial \mathcal{J}_{any|bce}}{\partial z_j} = \frac{\partial \mathcal{J}_{bce}}{\partial z_j} + \alpha \frac{\partial \mathcal{J}_{any}}{\partial z_j}
\end{equation}

We now use the chain rule to derive the learning signal for each final layer neuron during backpropagation. The partial derivative of standard BCE loss term \(\mathcal{J}_{bce}\) with respect to class \(j\) neuron is the product of the partial derivative of loss with respect to class \(j\) activation output \(p_j\) and the partial derivative of \(p_j\) with respect to pre-activation neuron \(z_j\):
\begin{equation}
    \frac{\partial \mathcal{J}_{bce}}{\partial z_j} = 
    \frac{\partial \mathcal{J}}{\partial p_j}  \frac{\partial p_j}{\partial z_j} =
    \frac{p_j - y_j}{p_j(1 - p_j)} \frac{p_j}{1 - p_j} =
    p_j - y_j
\end{equation}

To derive the additional gradient term propagated from any class loss term \(\mathcal{J}_{any}\) on each neuron, we further simplify the any class probability calculation \(p_a\):
\begin{equation}
\nonumber
    p_a = \frac{1}{1 + \frac{\prod_{j=1}^{M}(1-p_j)^{w_j/\sum_{j}w_j}}{\prod_{j=1}^{M}p_j^{w_j/\sum_{j}w_j}}} \\
\end{equation}
\begin{equation}
\nonumber
    p_a = \frac{1}{1 + \prod_{j=1}^{M}\left(\frac{1 - p_j}{p_j}\right)^{w_j/\sum_{j}w_j}} 
\end{equation}
\begin{equation}
\nonumber
    p_a = \frac{1}{1 + \prod_{j=1}^{M}(e^{-z_j})^{w_j/\sum_{j}w_j}} 
\end{equation}
\begin{equation}
    p_a = \frac{1}{1 + e^{-z^*}} \quad \text{where} \quad
    z^* = \frac{\sum_{j}w_j z_j}{\sum_{j}w_j}
\end{equation}
This further simplifies the any class presence probability \(p_a\) to the \(\mathrm{sigmoid}\) of a synthesized raw score \(z^*\) which is the weighted mean of final layer raw scores \(\mathbf{Z} = (z_j)_{j=1}^M\). Therefore, the learning signal on this synthesized raw score \(z^*\) is: 
\begin{equation}
    \frac{\partial \mathcal{J}_{any}}{\partial z^*} = p_a - y_a
\end{equation}
Tracing this gradient to each output neuron \(z_j\) is done as follows, which also explains how the gradient in each neuron depends on the positivity of the instance and the presence of class \(j\):
\begin{equation}
    \frac{\partial \mathcal{J}_{any}}{\partial z_j} = \frac{w_j}{\sum_{j}w_j} (p_a - y_a) \quad
    = \begin{cases}
        \frac{1}{\sum_{j}w_j} (p_a - 1) \quad \text{if} \quad y_a = 1 \quad \text{and} \quad  y_j = 1   \\
        \frac{\lambda}{\sum_{j}w_j} (p_a - 1) \quad \text{if} \quad y_a = 1 \quad \text{and} \quad  y_j = 0 \\
        \frac{1}{M} p_a \quad \text{if} \quad y_a = 0
    \end{cases}
\end{equation}
When an instance is positive, each final layer neuron is encouraged to contribute towards keeping any class presence probability \(p_a\) close to 1. Neurons corresponding to the present classes receive a stronger influence due to higher weights, while neurons for absent classes receive a lower weight \(\lambda\). When the instance is negative, each neuron is forced to keep any class probability \(p_a\) close to zero.

\subsection{Geometric mean and likelihood surface}
\label{apx:likelihood}
\begin{figure}[ht]
    \centering
    \begin{subfigure}[t]{0.45\textwidth}
        \centering
        \includegraphics[width=\linewidth]{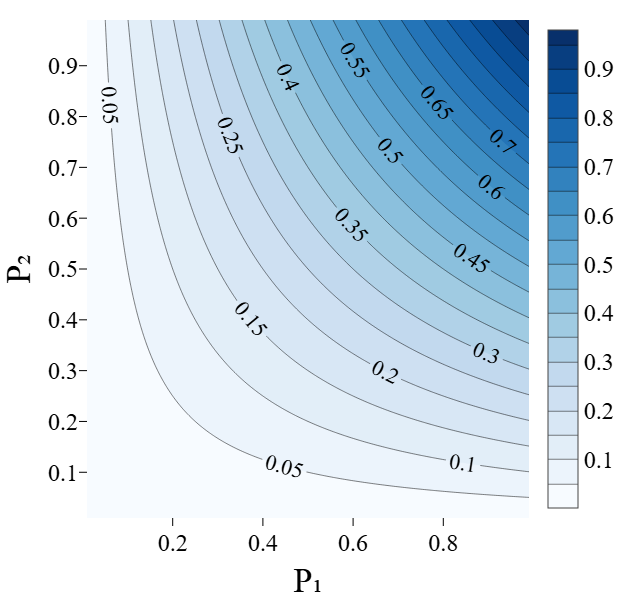}
        \caption{product \(p_1 p_2\)}
        \label{fig:prod_p1p2}
    \end{subfigure}
    \begin{subfigure}[t]{0.45\textwidth}
        \centering
        \includegraphics[width=\linewidth]{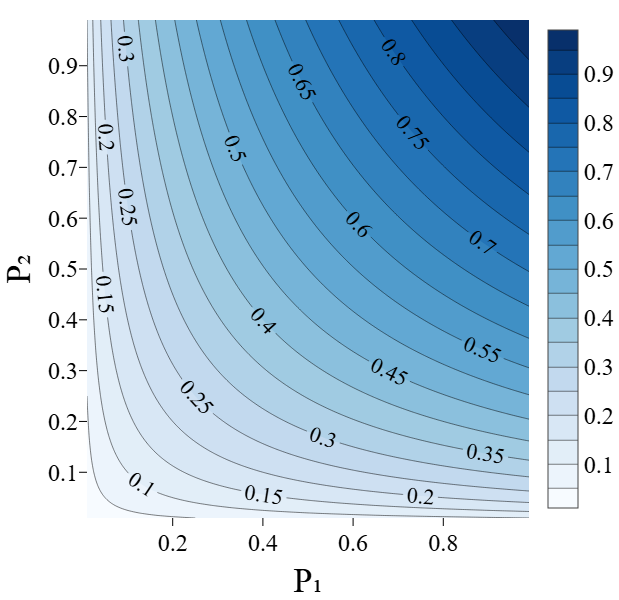}
        \caption{gmean \(\sqrt{p_1 p_2}\)}
        \label{fig:gmean_p1p2}
    \end{subfigure}
    \begin{subfigure}[t]{0.45\textwidth}
        \centering
        \includegraphics[width=\linewidth]{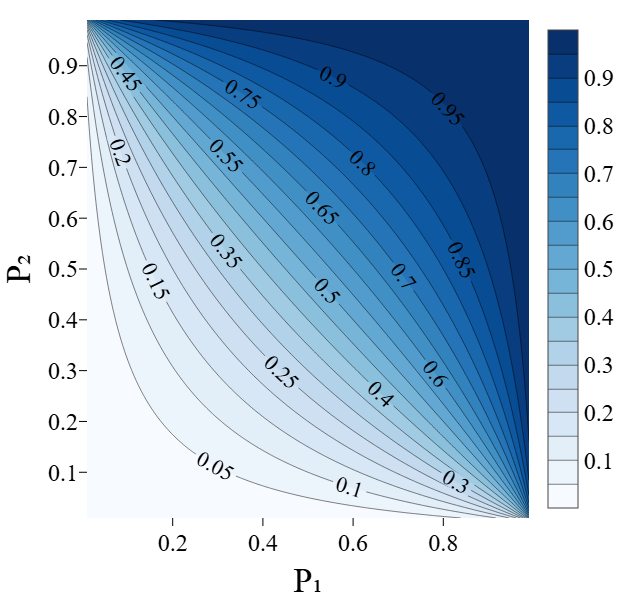}
        \caption{normalized product \(\frac{p_1 p_2}{p_1p_2 + (1-p1)(1-p_2)}\)}
        \label{fig:norm_prod_p1p2}
    \end{subfigure}
    \begin{subfigure}[t]{0.45\textwidth}
        \centering
        \includegraphics[width=\linewidth]{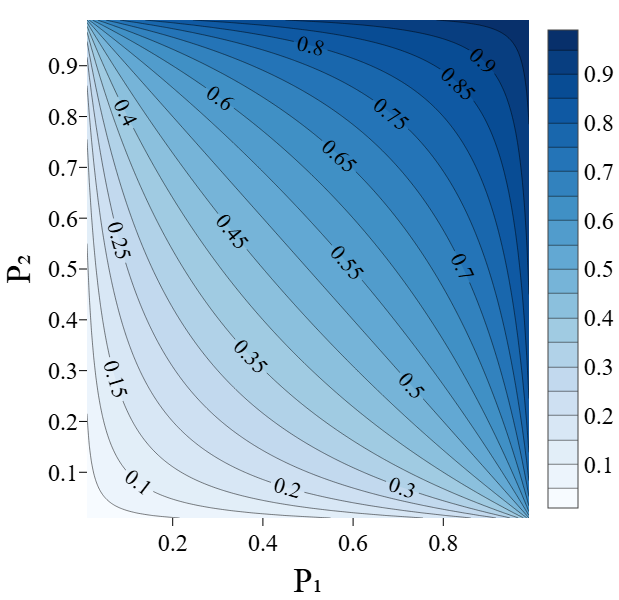}
        \caption{normalized gmean \(\frac{\sqrt{p_1 p_2}}{\sqrt{p_1 p_2} + \sqrt{(1-p_1)(1-p_2)}}\)}
        \label{fig:norm_gmean_p1p2}
    \end{subfigure}
    \caption{Surface plots of product and geometric mean between two probabilities, and their normalized surfaces. The geometric mean gives a better distribution on a robust scale for optimization.}
    \label{fig:gmean_vs_prod}
\end{figure}
In Equation~\eqref{eq:p_a}, the any-class probability is calculated with the normalized geometric mean of the predicted class probabilities. The geometric mean formulation fairly combines the individual probabilities, preserving the scale. To demonstrate this, Figure~\ref{fig:gmean_vs_prod} plots the product and geometric mean of two probabilities and their normalized variants. The plots show that the product gets saturated near $p_1=0$ and $p_2=0$, while the geometric mean remains sensitive to changes. This confirms that the geometric mean is a better formulation for likelihood.

\begin{figure}[ht]
    \centering
    \begin{subfigure}[b]{0.32\textwidth}
        \centering
        \includegraphics[width=\linewidth]{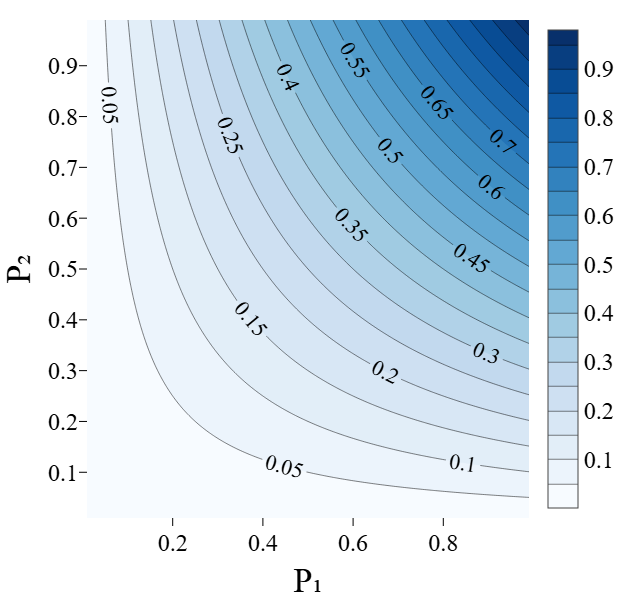}
        \caption{BCE \(y_1=1\) \(y_2=1\)}
        \label{fig:bce_like_11}
    \end{subfigure}
    \begin{subfigure}[b]{0.32\textwidth}
        \centering
        \includegraphics[width=\linewidth]{any_class_like_11_gmean.png}
        \caption{Any class \(y_1=1\) \(y_2=1\)}
        \label{fig:any_class_like_11}
    \end{subfigure}
    \begin{subfigure}[b]{0.32\textwidth}
        \centering
        \includegraphics[width=\linewidth]{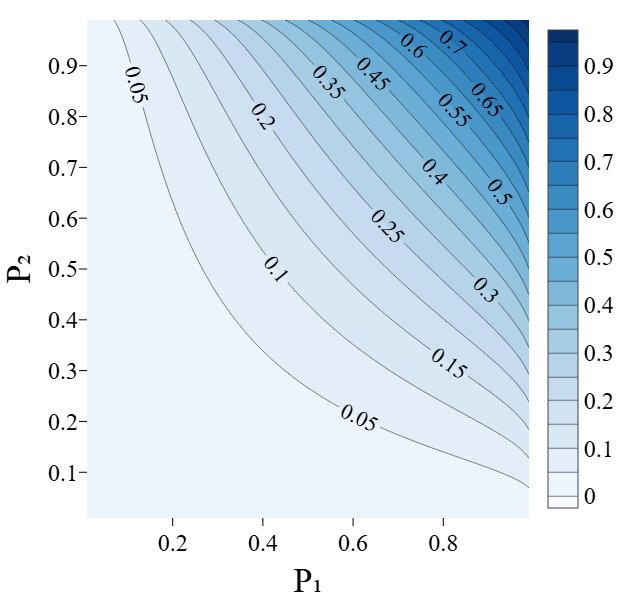}
        \caption{Redesigned BCE \(y_1=1\) \(y_2=1\)}
        \label{fig:redesigned_bce_like_11}
    \end{subfigure}
    \begin{subfigure}[b]{0.32\textwidth}
        \centering
        \includegraphics[width=\linewidth]{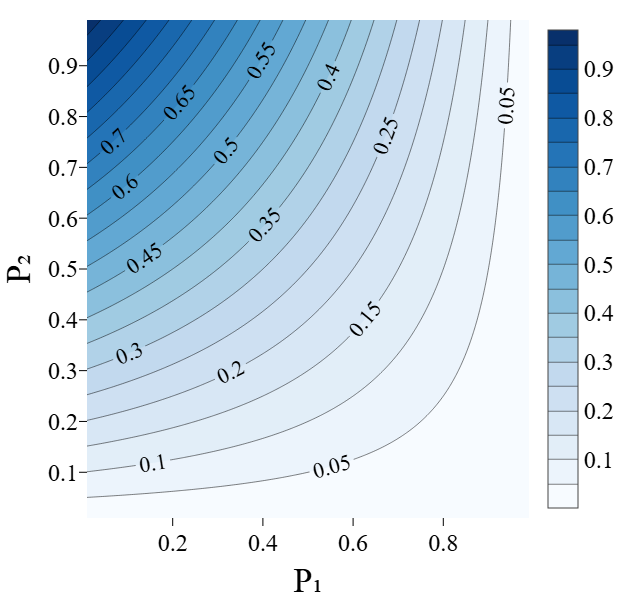}
        \caption{BCE \(y_1=0\) \(y_2=1\)}
        \label{fig:bce_like_01}
    \end{subfigure}
    \begin{subfigure}[b]{0.32\textwidth}
        \centering
        \includegraphics[width=\linewidth]{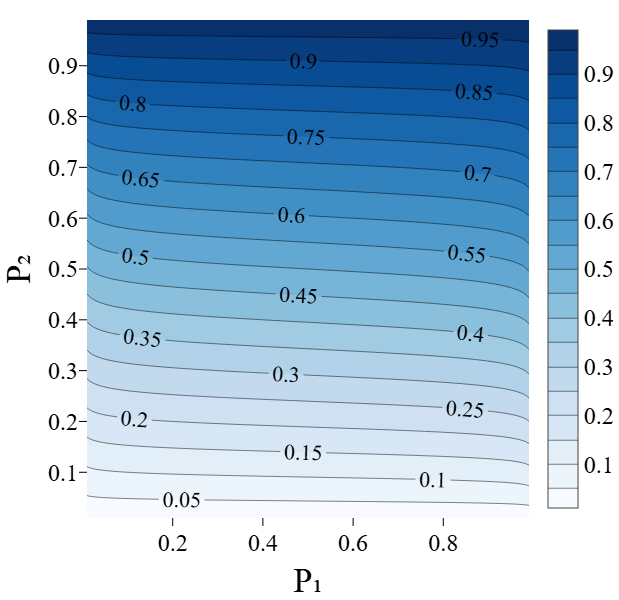}
        \caption{Any class \(y_1=0\) \(y_2=1\)}
        \label{fig:any_class_like_01}
    \end{subfigure}
    \begin{subfigure}[b]{0.32\textwidth}
        \centering
        \includegraphics[width=\linewidth]{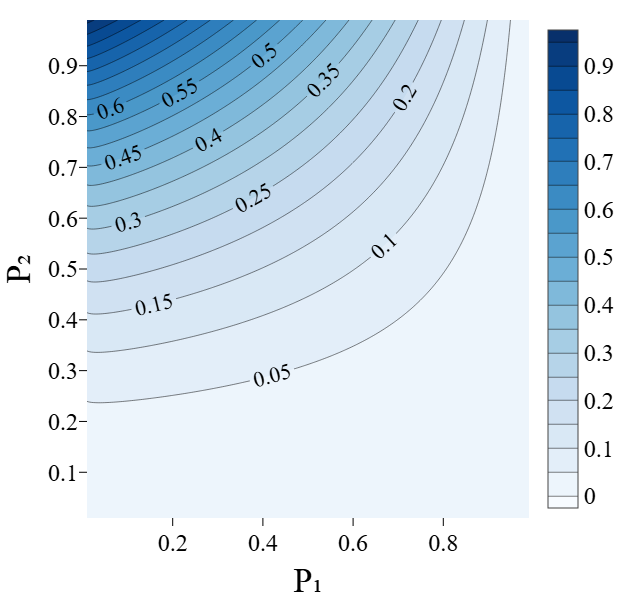}
        \caption{Redesigned BCE \(y_1=0\) \(y_2=1\)}
        \label{fig:redesigned_bce_like_01}
    \end{subfigure}
    \caption{Likelihood surface plots for two probabilities, the standard BCE likelihood, our any class likelihood and redesigned BCE likelihood. The first row depicts the case of both targets being 1, and in the second case, only one target is 1 and the any class likelihood is \(\lambda=0.05\)}
    \label{fig:likelihood_surface}
\end{figure}

Additionally, Figure~\ref{fig:likelihood_surface} presents the contour plots of BCE likelihood, the any-class likelihood, and the redesigned BCE likelihood for the two-class scenario. The first row of plots shows the scenario in which both classes are present (\(y_1=1\) and \(y_2=1\)), while the second row depicts the case where only one class is present (\(y_1=0\) and \(y_2=1\)). The two axes represent predicted probabilities \(p_1 \) and \(p_2\). Here, \(\lambda\) is set to 0.05 for any-class presence likelihood. In the two-class scenario, adding any-class likelihood (Figure~\ref{fig:any_class_like_11}) to BCE (Figure~\ref{fig:bce_like_11}) results in the redesigned loss (Figure~\ref{fig:redesigned_bce_like_11}) pushing the decision boundary towards 1-1 value (close to the label), making the redesigned loss a better option. Similarly, for the one-class scenario (the second row of plots), the redesign loss pushes the likelihood surface towards the true value of 0-1 with some skew caused by the $\lambda$ (Figure~\ref{fig:redesigned_bce_like_01}).

\subsection{Label distributions and COCO dataset class filtering}
\label{apx:coco-filtering}
\begin{table*}[ht]
\centering
\caption{Summary of dataset distributions}
\label{tab:dataset-distribution}
\begin{tabular}{llrrr}
\toprule
\textbf{Dataset} & \textbf{Split} & \textbf{Number of Samples} & \textbf{Any-Class} & \textbf{Negative} \\
\midrule
\multirow{4}{*}{Chest X-ray} 
  & Train      & 78,566   & 36,356   & 42,210 \\
  & Validation & 17,063   & 7,821    & 9,242  \\
  & Test       & 16,491   & 7,582    & 8,909  \\
  & Total      & 112,120  & 51,759   & 60,361 \\
\midrule
\multirow{4}{*}{COCO} 
  & Train      & 86,300   & 45,658   & 40,642 \\
  & Validation & 18,493   & 9,784    & 8,709  \\
  & Test       & 18,494   & 9,784    & 8,710  \\
  & Total      & 123,287  & 65,226   & 58,061 \\
\midrule
\multirow{2}{*}{Sewer} 
  & Train      & 1,040,129 & 487,309  & 552,820 \\
  & Validation & 130,046   & 61,365   & 68,681  \\
\bottomrule
\end{tabular}
\end{table*}

We evaluate our models on three diverse datasets spanning infrastructure, general object, and medical domains: SewerML, COCO, and Chest X-ray. Each dataset is split into training, validation, and testing subsets, with a mix of samples containing at least one class label (“Any-Class”) and samples with no target classes (“Negative”). Table~\ref{tab:dataset-distribution} summarizes the distribution statistics across all splits for each dataset.

\begin{figure}[ht]
    \centering
    \begin{subfigure}[t]{0.45\textwidth}
        \centering
        \begin{tikzpicture}
\begin{axis}[
    ybar,
    bar width=4pt,
    width=\textwidth,
    height=6cm, 
    xlabel={Class},
    ylabel={Number of Annotations},
    symbolic x coords={
        person,car,chair,book,bottle,cup,dining table,bowl,traffic light,
        handbag,umbrella,bird,boat,truck,
        bench,sheep,banana,kite,motorcycle,
    },
    xtick=data,
    xticklabel style={rotate=90,  anchor=east, font=\tiny},
    enlarge x limits=0.01,
    ymin=0,
    axis x line*=bottom,
    axis y line*=left,
    every axis plot/.append style={draw=blue, fill=blue!30}
]
\addplot coordinates {
    (person,262465)(car,43867)(chair,38491)(book,24715)(bottle,24342)
    (cup,20650)(dining table,15714)(bowl,14358)(traffic light,12884)
    (handbag,12354)(umbrella,11431)(bird,10806)(boat,10759)(truck,9973)
    (bench,9838)(sheep,9509)(banana,9458)(kite,9076)(motorcycle,8725)
};
\end{axis}
\end{tikzpicture}
        \caption{COCO: 20 most occurring class distribution.}
        \label{fig:class-distribution}
    \end{subfigure}
    \begin{subfigure}[t]{0.45\textwidth}
        \centering
        \begin{tikzpicture}
\begin{axis}[
    ybar,
    bar width=8pt,
    width=\textwidth,
    height=6cm,
    xlabel={Category},
    ylabel={Number of Annotations},
    symbolic x coords={
        Person, Vehicles, Kitchenware, Furniture, Food, Animals,
        Sports, Objects, Accessories, Electronics, Road Objects, Appliances
    },
    xtick=data,
    xticklabel style={rotate=90, anchor=east, font=\tiny},
    ymin=0,
    enlarge x limits=0.05,
    axis x line*=bottom,
    axis y line*=left,
    every axis plot/.append style={draw=blue, fill=blue!30}
]
\addplot coordinates {
    (Person, 262465)
    (Vehicles, 96212)
    (Kitchenware, 86677)
    (Furniture, 76985)
    (Food, 63512)
    (Animals, 62566)
    (Sports, 50940)
    (Objects, 46088)
    (Accessories, 45193)
    (Electronics, 28029)
    (Road Objects, 27855)
    (Appliances, 13479)
};
\end{axis}
\end{tikzpicture}
        \caption{COCO: grouped category distribution.}
        \label{fig:category-distribution}
    \end{subfigure}
    \begin{subfigure}[t]{\textwidth}
        \centering

\begin{tikzpicture}
\begin{axis}[
    width=8cm,
    height=8cm,
    xlabel={Category Group},
    ylabel={Category Group},
    xlabel style={font=\small},
    ylabel style={font=\small},
    colormap={custom}{color(0)=(white) color(0.15)=(white!95!blue) color(0.5)=(blue!50!white) color(1)=(blue)},
    point meta min=0,
    point meta max=24000,
    xtick={0,...,11},
    ytick={0,...,11},
    xticklabels={Accessories, Animals, Appliances, Electronics, Food, Furniture, Kitchenware, Objects, Person, Road Objects, Sports, Vehicles},
    yticklabels={Accessories, Animals, Appliances, Electronics, Food, Furniture, Kitchenware, Objects, Person, Road Objects, Sports, Vehicles},
    xticklabel style={rotate=90, anchor=east, font=\scriptsize, text=black!70},
    yticklabel style={font=\scriptsize, text=black!70},
    tick align=outside,
    grid=none,
    xmin=-0.5, xmax=11.5,
    ymin=-0.5, ymax=11.5,
    axis on top,
    cell picture=true,
    point meta=explicit,
]
\addplot [
    matrix plot,
    mesh/cols=12,
    point meta=explicit,
] table [meta=z] {heatmap_data.dat};
\end{axis}
\end{tikzpicture}
        \caption{Category-level co-occurrence heatmap.}
        \label{fig:heatmap-categories}
    \end{subfigure}
    \caption{COCO: 20 most occurring class distribution, abstract category distribution, and category co-occurrence heatmap. Person is the most frequent class and has the most co-occurrences with other classes.}
    \label{fig:coco_distributions}
\end{figure}
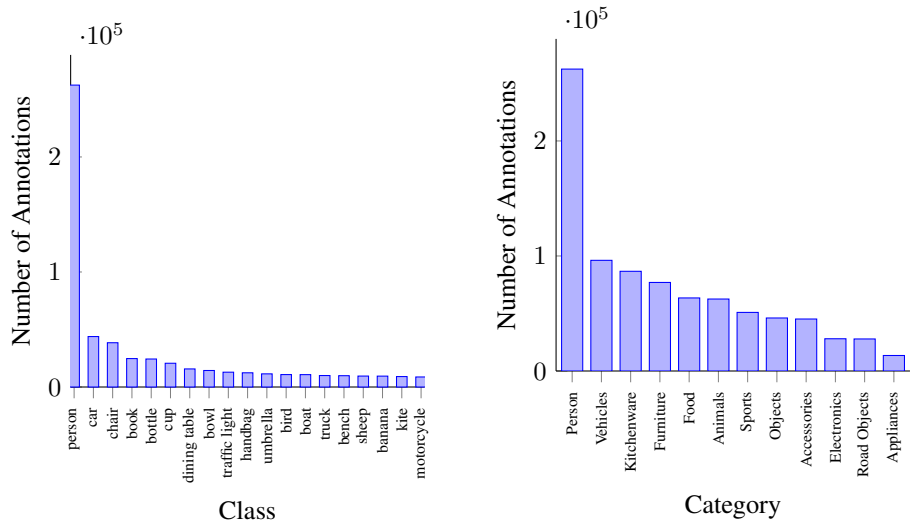
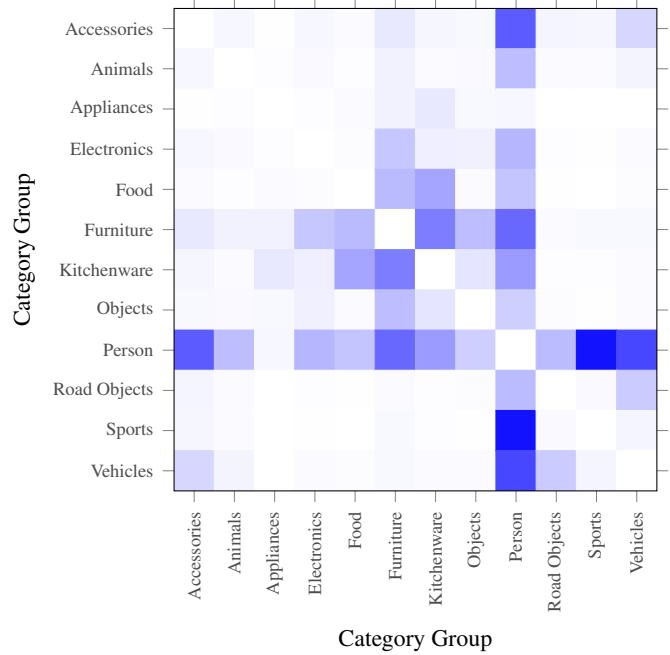

In its original form, the COCO dataset does not include negative images. This poses challenges for training multi-label models in contexts where object absence is meaningful. To create a dataset with negative classes, we devised a structured class-removing strategy based on statistical analysis of the dataset. By analyzing the raw annotation frequency of the 20 most occurring object classes in the dataset, as shown in Figure~\ref{fig:class-distribution}, it is evident that the \textit{person} class dominates the dataset with over 260,000 annotations. Thus, the \textit{person} class was the first to be removed. The classes were then grouped into related categories such as \textit{Furniture}, \textit{Kitchenware}, \textit{Vehicles}, and similar. The grouped distribution, shown in Figure~\ref{fig:category-distribution}, confirmed that the \textit{People} group remained the most dominant, supporting the decision to remove it.

To identify which categories are highly correlated with the \textit{person} class, category-level co-occurrence heatmaps were examined, as seen in Figure~\ref{fig:heatmap-categories}. These visualizations revealed strong co-occurrence of \textit{person} with categories such as \textit{Accessories}, \textit{Sports}, and \textit{Vehicles}. Based on these findings, we applied a filtering process to refine the label set. Following the removal of \textit{person} class, classes that strongly correlate with \textit{person} were removed, particularly those related to accessories, sports, and vehicles. To minimize semantic overlap and ambiguity, we also removed highly similar classes such as \textit{bicycle} and \textit{motorcycle} or \textit{cup} and \textit{wine glass}.

\begin{table}[t]
\centering
\caption{COCO: remaining classes}
\label{tab:remaining-classes}
\begin{tabular}{llllll}
\toprule
\textbf{Classes} & & & & & \\
\midrule
boat & bird & cat & dog & bottle & fork \\
knife & spoon & banana & apple & sandwich & orange \\
broccoli & carrot & hot dog & pizza & donut & cake \\
chair & couch & bed & dining table & toilet & tv \\
laptop & mouse & remote & keyboard & cell phone & microwave \\
oven & toaster & sink & refrigerator & book & clock \\
vase & scissors & teddy bear & hair drier & toothbrush & \\
\bottomrule
\end{tabular}
\end{table}

The final list of remaining classes is shown in Table~\ref{tab:remaining-classes}. This systematic pruning allowed us to introduce class-absent examples, which are critical for evaluating models that can distinguish between the presence or absence of relevant objects in a scene.

\subsection{Additional experiments}
\label{apx:experiments}
\begin{table}[ht]
\setlength{\tabcolsep}{4pt}
\caption{TresNet-L performance with varying \(\lambda\) values for the ChestX-ray14 dataset with BCE loss, COCO with extreme negatives with BCE Loss, and for the SewerML dataset with focal loss}
\label{tab:apx_lambda}
\centering
\begin{tabular}{lllllllllll}
\toprule
\multirow{2}{*}{Dataset} &  \multirow{2}{*}{Metric} & \multirow{2}{*}{\(\mathcal{J}^{cb}_{bce}\)} & \multicolumn{8}{c}{ \(\mathcal{J}^{cb}_{any|bce}\) with varying \(\lambda\) }  \\
\cmidrule(r){4-11}
& & &  0  & 0.01 & 0.02  & 0.05  & 0.1  & 0.2  & 0.5 & 1  \\
\midrule
ChestX-ray14 & F1       & 13.98       & 19.25 & 18.7 & \textbf{19.99} & 19.04 & 17.98 & 18.26 & 16.58 & 16.35 \\
             & F2       & 12.77       & 19.53 & 19.02 & \textbf{20.43} & 19.28 & 18.38 & 18.46 & 16.3 & 15.81\\
             & AP       & 17.55       & 17.79 & 17.69 & 18.53 & 18.33 & 18.44 & \textbf{18.59} & 18.5 & 18.25 \\
             & F1-Neg   & 72.83       & 71.99 & 72.09 & 71.95 & 72.72 & 72.62 & 72.98 & \textbf{73.17} & 73.16\\   
\midrule
COCO  & F1       & 55.58 & 55.54 & 56.15 & 56.63 & 57.46 & 56.31 & \textbf{57.91} & 56.94 & 57.23 \\
Extreme Negative             & F2     &  52.84 & 55.79 & 57.61 & 58.20 & 59.62 & 59.38 & \textbf{60.62} & 58.11 & 57.37\\
             & AP       & 58.12 & 56.36 & 58.14 & 59.14 & 61.00 & 59.72 & \textbf{61.58} & 60.68 & 60.63\\
             & F1-Neg   & 92.75 & 91.04 & 90.69 & 90.73 & 91.23 & 91.14 & 92.04 & 93.06 & \textbf{93.18}\\ 
\toprule
\multirow{2}{*}{Dataset} &  \multirow{2}{*}{Metric} & \multirow{2}{*}{\(\mathcal{J}^{cb}_{focal}\)} & \multicolumn{8}{c}{ \(\mathcal{J}^{cb}_{any|focal}\) with varying \(\lambda\) }  \\
\cmidrule(r){4-11}
& & &  0  & 0.01 & 0.02  & 0.05  & 0.1  & 0.2  & 0.5 & 1  \\
\midrule
SewerML & F1 & 59.73 & 63.06 & 62.75 & \textbf{63.33} & 63.06 & 63.06 & 63.04 & 61.43 & 60.72 \\
        & F2 & 56.61  & 62.28 & 62.37 & \textbf{63.05} & 62.71 & 62.80 & 62.50 & 60.00 & 58.81 \\
        & AP & 62.71 & 65.44 & 65.28 & \textbf{65.82} & 65.34 & 65.53 & 65.37 & 63.68 & 62.55 \\
        & F1-Neg & 91.21  & 91.75  & 91.51  & 91.59  & 91.53  & 91.34  & \textbf{91.89}  & 91.75  & 91.56\\    
\bottomrule
\end{tabular}
\end{table}

\begin{figure}[ht]
    \centering
    \begin{subfigure}[b]{0.49\textwidth}
        \centering
        \pgfplotstableread{
lambda   F1     F2     AP     Recall  F1-Any  F1-Neg
0.005   19.25	19.53	17.79	20.17	61.03	71.99
0.01    18.7	19.02	17.69	19.67	61.6	72.09
0.02    19.99	20.43	18.53	21.21	61.86	71.95
0.05    19.04	19.28	18.33	19.84	60.59	72.72
0.1     17.98	18.38	18.44	19.33	59.2	72.62
0.2     18.26	18.46	18.59	19.16	58.93	72.98
0.5     16.58	16.3	18.5	16.67	54.89	73.17
1       16.35	15.81	18.25	15.85	53.48	73.16
}\datatable

\begin{tikzpicture}
\begin{loglogaxis}[
    xlabel={\scriptsize{$\lambda$ (in log scale)}},
    ylabel={\scriptsize{Score (\%)}},
    log ticks with fixed point,
    xtick={0.005, 0.01, 0.02, 0.05, 0.1, 0.2, 0.5, 1},
    xticklabels={0, 0.01, 0.02, 0.05, 0.1, 0.2, 0.5, 1},  
    xticklabel style={font=\scriptsize},
    ymin=9,
    ytick={10, 15, 20, 25, 30},
    yticklabels={10, 15, 20, 70, 75},  
    yticklabel style={font=\scriptsize, /pgf/number format/fixed},
    legend style={
        at={(0.5,-0.25)},
        anchor=north,
        draw=none,
        fill=none,
        font=\scriptsize,
        legend columns=5,
        /tikz/every even column/.style={column sep=0.3cm}
    },
    width=\linewidth,
    height=5cm,
    grid=major,
    xmajorgrids=false,
    ymajorgrids=true,
    tick style={black},
    every axis label/.append style={font=\small},
    label style={font=\small},
    title style={font=\footnotesize},
    legend cell align=left
]

\addplot+[color=blue, mark=o, thick, forget plot] table[x=lambda, y=F1] from \datatable; 
\addplot+[color=orange, mark=square, thick, forget plot] table[x=lambda, y=F2] from \datatable; 
\addplot+[color=purple, mark=triangle, thick, forget plot] table[x=lambda, y=AP] from \datatable;
\addplot+[color=red!70!black, mark=star, thick, forget plot] table[x=lambda, y expr=\thisrow{F1-Neg} - 45] from \datatable;

\addplot+[color=blue, mark=o, only marks, thick] coordinates {(0.0025, 13.98)};
\addplot+[color=orange, mark=square, only marks, thick] coordinates {(0.0025, 12.77)};
\addplot+[color=purple, mark=triangle, only marks, thick] coordinates {(0.0025, 17.55)};
\addplot+[color=red!70!black, mark=star, only marks, thick] coordinates {(0.0025, 72.83 - 45)};

\path[fill=gray!30, fill opacity=0.5]
    (axis cs:0.0016, 5) rectangle (axis cs:0.0037, 40);

\node[font=\tiny\itshape] at (axis cs:0.0025, 10.5) {$\mathcal{J}^{cb}_{bce}$};
\node[font=\tiny\itshape] at (axis cs:0.01, 10.5) {$\mathcal{J}^{cb}_{any|bce}$};

\draw[black, thick, dashed] 
    (axis cs:0.0015, 22.5) -- (axis cs:1,22.5);

\addlegendimage{only marks, mark=o, color=blue}
\addlegendentry{F1}
\addlegendimage{only marks, mark=square, color=orange}
\addlegendentry{F2}
\addlegendimage{only marks, mark=triangle, color=purple}
\addlegendentry{AP}
\addlegendimage{only marks, mark=star, color=red!70!black}
\addlegendentry{F1-Neg}

\end{loglogaxis}
\end{tikzpicture}
        \caption{ChestX-ray14 - BCE variant}
        \label{fig:lambda-chest-tresnet-bce}
    \end{subfigure}
    \begin{subfigure}[b]{0.49\textwidth}
        \centering
        \pgfplotstableread{
lambda   F1     F2     AP     F1-Neg
0.005   55.54   55.79   56.36   91.04
0.01    56.15   57.61   58.14   90.69
0.02    56.63   58.2    59.14   90.73
0.05    57.46   59.62   61.00   91.23
0.1     56.31   59.38   59.72   91.14
0.2     57.91   60.62   61.58   92.04
0.5     56.94   58.11   60.68   93.06
1       57.23   57.37   60.63   93.18
}\datatable

\begin{tikzpicture}
\begin{loglogaxis}[
    xlabel={\scriptsize{$\lambda$ (in log scale)}},
    ylabel={\scriptsize{Score (\%)}},
    log ticks with fixed point,
    xtick={0.005, 0.01, 0.02, 0.05, 0.1, 0.2, 0.5, 1},
    xticklabels={0, 0.01, 0.02, 0.05, 0.1, 0.2, 0.5, 1},  
    xticklabel style={font=\scriptsize},
    ymin= 48,
    ytick={50, 55, 60, 65, 70, 75},
    yticklabels={50, 55, 60, 65, 90, 95},  
    yticklabel style={font=\scriptsize, /pgf/number format/fixed},
    legend style={
        at={(0.5,-0.25)},
        anchor=north,
        draw=none,
        fill=none,
        font=\scriptsize,
        legend columns=5,
        /tikz/every even column/.style={column sep=0.3cm}
    },
    width=\linewidth,
    height=5cm,
    grid=major,
    xmajorgrids=false,
    ymajorgrids=true,
    tick style={black},
    every axis label/.append style={font=\small},
    label style={font=\small},
    title style={font=\footnotesize},
    legend cell align=left
]

\addplot+[color=blue, mark=o, thick, forget plot] table[x=lambda, y=F1] from \datatable; 
\addplot+[color=orange, mark=square, thick, forget plot] table[x=lambda, y=F2] from \datatable; 
\addplot+[color=purple, mark=triangle, thick, forget plot] table[x=lambda, y=AP] from \datatable;
\addplot+[color=red!70!black, mark=star, thick, forget plot] table[x=lambda, y expr=\thisrow{F1-Neg}-20] from \datatable;

\addplot+[color=blue, mark=o, only marks, thick] coordinates {(0.0025, 55.58)};
\addplot+[color=orange, mark=square, only marks, thick] coordinates {(0.0025, 52.84)};
\addplot+[color=purple, mark=triangle, only marks, thick] coordinates {(0.0025, 58.12)};
\addplot+[color=red!70!black, mark=star, only marks, thick] coordinates {(0.0025, 92.75 - 20)};

\path[fill=gray!30, fill opacity=0.5]
    (axis cs:0.0016, 40) rectangle (axis cs:0.0037, 90);

\node[font=\tiny\itshape] at (axis cs:0.0025, 50) {$\mathcal{J}^{cb}_{bce}$};
\node[font=\tiny\itshape] at (axis cs:0.01, 50) {$\mathcal{J}^{cb}_{any|bce}$};

\draw[black, thick, dashed] 
    (axis cs:0.0015, 67) -- (axis cs:1,67);

\addlegendimage{only marks, mark=o, color=blue}
\addlegendentry{F1}
\addlegendimage{only marks, mark=square, color=orange}
\addlegendentry{F2}
\addlegendimage{only marks, mark=triangle, color=purple}
\addlegendentry{AP}
\addlegendimage{only marks, mark=star, color=red!70!black}
\addlegendentry{F1-Neg}

\end{loglogaxis}
\end{tikzpicture}
        \caption{COCO extreme negatives - BCE variant}
        \label{fig:lambda-coco_extreme_tresnet_bce}
    \end{subfigure}
    \begin{subfigure}[b]{0.49\textwidth}
        \centering
        \pgfplotstableread{
lambda	F1	      F2	AP	    Recall	F1-Any	F1-Neg
0.005	    63.06	62.28	65.44	62	    91.25	91.75
0.01	62.75	62.37	65.28	62.35	91.13	91.51
0.02	63.33	63.05	65.82	63.09	91.23	91.59
0.05	63.06	62.71	65.34	62.71	91.16	91.53
0.1	    63.06	62.8	65.53	62.87	91.34	91.34
0.2	    63.04	62.5	65.37	62.4	91.38	91.89
0.5	    61.43	60	    63.68	59.31	91.02	91.75
1	    60.72	58.81	62.55	57.83	90.7	91.56
}\datatable

\begin{tikzpicture}
\begin{loglogaxis}[
    xlabel={\scriptsize{$\lambda$ (in log scale)}},
    ylabel={\scriptsize{Score (\%)}},
    log ticks with fixed point,
    xtick={0.005, 0.01, 0.02, 0.05, 0.1, 0.2, 0.5, 1},
    xticklabels={0, 0.01, 0.02, 0.05, 0.1, 0.2, 0.5, 1},  
    xticklabel style={font=\scriptsize},
    ymin=52,
    ytick={55, 60, 65, 70, 75},
    yticklabels={55, 60, 65, 90, 95},  
    yticklabel style={font=\scriptsize, /pgf/number format/fixed},
    legend style={
        at={(0.5,-0.25)},
        anchor=north,
        draw=none,
        fill=none,
        font=\scriptsize,
        legend columns=5,
        /tikz/every even column/.style={column sep=0.3cm}
    },
    width=\linewidth,
    height=5cm,
    grid=major,
    xmajorgrids=false,
    ymajorgrids=true,
    tick style={black},
    every axis label/.append style={font=\small},
    label style={font=\small},
    title style={font=\footnotesize},
    legend cell align=left
]

\addplot+[color=blue, mark=o, thick, forget plot] table[x=lambda, y=F1] from \datatable; 
\addplot+[color=orange, mark=square, thick, forget plot] table[x=lambda, y=F2] from \datatable; 
\addplot+[color=purple, mark=triangle, thick, forget plot] table[x=lambda, y=AP] from \datatable; 
\addplot+[color=red!70!black, mark=star, thick, forget plot] 
    table[x=lambda, y expr=\thisrow{F1-Neg}-20] from \datatable;

\addplot+[color=blue, mark=o, only marks, thick] coordinates {(0.0025, 59.73)};
\addplot+[color=orange, mark=square, only marks, thick] coordinates {(0.0025, 56.61)};
\addplot+[color=purple, mark=triangle, only marks, thick] coordinates {(0.0025, 62.71)};
\addplot+[color=red!70!black, mark=star, only marks, thick] coordinates {(0.0025, 91.21 - 20)};

\path[fill=gray!30, fill opacity=0.5]
    (axis cs:0.0016, 50) rectangle (axis cs:0.0037, 90);

\node[font=\tiny\itshape] at (axis cs:0.0025, 54) {$\mathcal{J}^{cb}_{focal}$};
\node[font=\tiny\itshape] at (axis cs:0.012, 54) {$\mathcal{J}^{cb}_{any|focal}$};

\draw[black, thick, dashed] 
    (axis cs:0.0015, 68.5) -- (axis cs:1,68.5);

\addlegendimage{only marks, mark=o, color=blue}
\addlegendentry{F1}
\addlegendimage{only marks, mark=square, color=orange}
\addlegendentry{F2}
\addlegendimage{only marks, mark=triangle, color=purple}
\addlegendentry{AP}
\addlegendimage{only marks, mark=star, color=red!70!black}
\addlegendentry{F1-Neg}

\end{loglogaxis}
\end{tikzpicture}
        \caption{SewerML - focal variant}
        \label{fig:lambda-sewer-tresnet-focal}
    \end{subfigure}
    \caption{Performance across varying $\lambda$ values for TresNet-L network for setting in Table~\ref{tab:apx_lambda}}
    \label{fig:apx_lambda}
\end{figure}
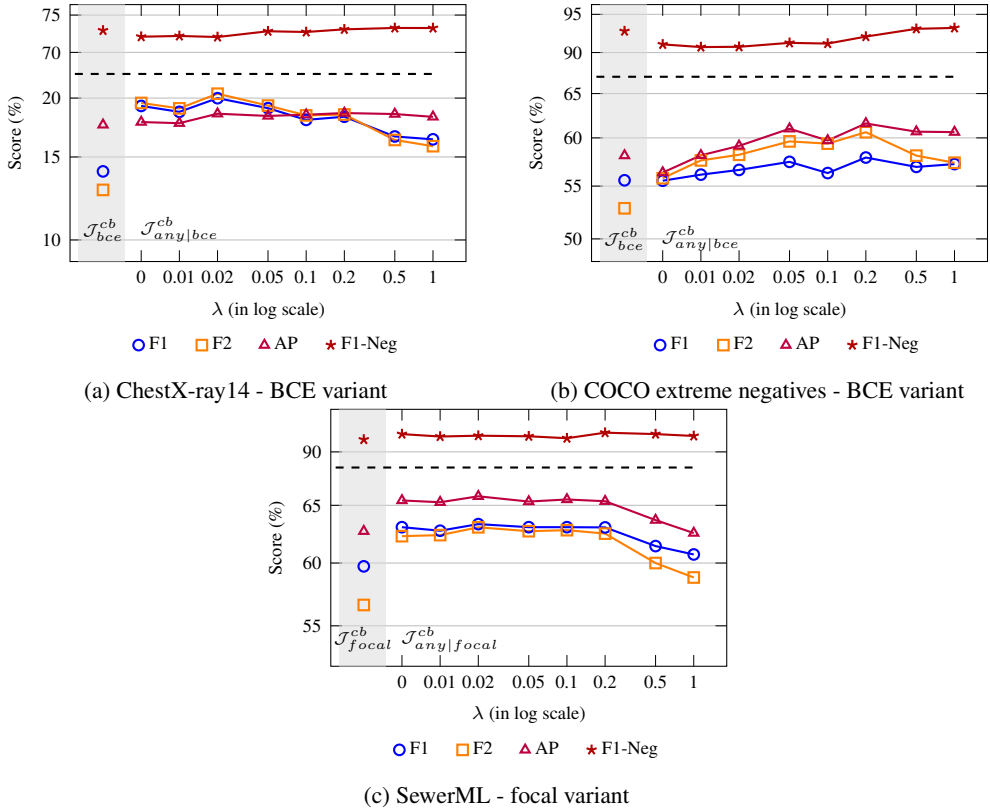

The main ablation study is presented in Subsection~\ref{ss:ablation_lambda}, while in this section, we report additional ablation experiments on the effect of varying \(\lambda\). Table~\ref{tab:apx_lambda} and Figure~\ref{fig:apx_lambda} show the impact of \(\lambda\) on ChestX-ray14 with BCE Loss, and SewerML with Focal Loss. We also built another custom COCO label set with only animal classes, leaving 80\% negative instances. While other cases presented here agree with the ablation study presented in Subsection~\ref{ss:ablation_lambda}, the extreme negative COCO dataset shows the best MLC performance for higher \(\lambda\) (\(\lambda=0.2\)). This highlights the need for careful tuning of \(\lambda\) for the optimum contrasting of MLC performance from negative data, depending on the occurrence of negative data in the respective task.

\end{document}